\documentclass[mnsc,nonblindrev]{informs3_hide}
\OneAndAHalfSpacedXI 

\usepackage[utf8]{inputenc} 
\usepackage[english]{babel}
\usepackage[autostyle, english = american]{csquotes}
\MakeOuterQuote{"}
\usepackage[colorlinks,linkcolor=blue, citecolor=blue]{hyperref}
\usepackage[normalem]{ulem}

\usepackage{cleveref}
\crefname{subsection}{section}{subsections}
\usepackage{eqnarray}

\usepackage[dvipsnames]{xcolor}

\usepackage{xparse}

\NewDocumentEnvironment{myproof}{o}
{\IfNoValueTF{#1}{\paragraph{{Proof.} }} {\paragraph{{#1.} }} }
{\hfill$\Halmos$}

\usepackage{natbib,bbm,multirow,multicol}
 \bibpunct[, ]{(}{)}{,}{a}{}{,}%
 \def\bibfont{\small}%
\TheoremsNumberedThrough     
\ECRepeatTheorems

\EquationsNumberedThrough    

\usepackage[T1]{fontenc}    
\usepackage{hyperref}       
\usepackage{url}            
\usepackage{booktabs}       
\usepackage{amsfonts}       
\usepackage{nicefrac}       
\usepackage{microtype}      
\usepackage{algorithm}
\usepackage[noend]{algorithmic}
\usepackage{amsmath, bbm, enumitem, xparse, bm, lipsum}
\usepackage{amssymb}
\usepackage{natbib,bbm,multirow,multicol}
\usepackage{graphicx}
\usepackage[inkscapelatex=false]{svg}
\usepackage{float}
\usepackage{subfigure}
\usepackage[normalem]{ulem}               
\usepackage{anyfontsize}
\begin{document}


%

\TITLE{Reinforcement Learning with Intrinsically Motivated Feedback Graph for Lost-sales Inventory Control}

\ARTICLEAUTHORS{%
\AUTHOR{LIU Zifan}
\AFF{ 
  Department of Electronic and Computer Engineering,
  The Hong Kong University of Science and Technology,
  \texttt{zliuft@connect.ust.hk}
}
\AUTHOR{LI Xinran}
\AFF{ 
  Department of Electronic and Computer Engineering,
  The Hong Kong University of Science and Technology,
  \texttt{xinran.li@connect.ust.hk}
}
\AUTHOR{Chen Shibo}
\AFF{ 
  Department of Electronic and Computer Engineering,
  The Hong Kong University of Science and Technology,
  \texttt{eeshibochen@ust.hk}
}
\AUTHOR{LI Gen}
\AFF{ 
  Department of Statistics,
  The Chinese University of Hong Kong,\\
  \texttt{genli@cuhk.edu.hk}
}
\AUTHOR{JIANG Jiashuo}
\AFF{ 
  Department of Industrial Engineering and Decision Analytics,
  The Hong Kong University of Science and Technology,
  \texttt{jsjiang@ust.hk}
}
\AUTHOR{Jun Zhang}
\AFF{ 
  Department of Electronic and Computer Engineering,
  The Hong Kong University of Science and Technology,\\
  \texttt{eejzhang@ust.hk}
}

}

\ABSTRACT{
Reinforcement learning (RL) has proven to be well-performed and versatile in inventory control (IC). However, further improvement of RL algorithms in the IC domain is impeded by two limitations of online experience. First, online experience is expensive to acquire in real-world applications. With the low sample efficiency nature of RL algorithms, it would take extensive time to collect enough data and train the RL policy to convergence. Second, online experience may not reflect the true demand due to the lost-sales phenomenon typical in IC, which makes the learning process more challenging. To address the above challenges, we propose a training framework that combines reinforcement learning with feedback graph (RLFG) and intrinsically motivated exploration (IME) to boost sample efficiency. In particular, we first leverage the MDP structure inherent in lost-sales IC problems and design the feedback graph (FG) tailored to lost-sales IC problems to generate abundant side experiences aiding in RL updates. Then we conduct a rigorous theoretical analysis of how the designed FG reduces the sample complexity of RL methods. Guided by these insights, we design an intrinsic reward to direct the RL agent to explore to the state-action space with more side experiences, further exploiting FG’s capability. Experimental results on single-item, multi-item, and multi-echelon environments demonstrate that our method greatly improves the sample efficiency of applying RL in IC. 
Our code is available at \url{https://github.com/Ziffer-byakuya/RLIMFG4IC}
}



\maketitle

\section{Introduction}\label{sec:intro}
Inventory control (IC) is a practical problem rooted in the real-world application of supply chain management (SCM). It mainly aims at minimizing operational costs while satisfying the demand of customers. The inherent complexity of modeling supply chain dynamics and demand uncertainty makes finding the optimal solutions for IC computationally intractable within reasonable time frames. To address this challenge, researchers \citep{zipkin2008old,bai2023asymptotic,xin2021understanding,chen2024learning} have designed heuristic methods based on specific model assumptions over the past few decades. However, these methods face limitations in practice. Firstly, they struggle with the curse of dimensionality \citep{goldberg2021survey}. As the lead time from order placement to reception increases, the problem size grows exponentially. Secondly, their reliance on specific assumptions may restrain their adaptability; they cannot easily generalize across diverse environmental settings. For example, most well-performed heuristic methods for single-item IC cannot be extended to multi-item or multi-echelon IC. These limitations call for a more adaptable approach to IC problems, for which data-driven control methods stand out as a promising alternative.

Reinforcement learning (RL) has shown promise in handling complex sequential decision-making problems. In particular, it offers several advantages for addressing the challenges of IC problems. Firstly, RL allows for the discovery of optimal policies without relying on strong problem-specific assumptions, enabling more generalizable solutions \citep{nian2020review}. Secondly, when combined with the deep neural network, Deep RL (DRL) can handle the high-dimensional state and action spaces typical of IC, thereby alleviating the curse of dimensionality \citep{dehaybe2024deep}. Early works have demonstrated the feasibility of RL in IC problems. For instance, \cite{oroojlooyjadid2022deep} showcases the ability of Deep Q-network (DQN) to discover near-optimal solutions for the widely recognized beer distribution game (a type of game for SCM simulation that illustrates the complexities and challenges of IC). \cite{gijsbrechts2022can} achieves performance comparable to that of heuristic methods using the A3C algorithm in IC problems. \cite{stranieri2023comparing} benchmarks various DRL methods such as A3C, PPO, and vanilla policy gradient (VPG) in IC problems. More recently, researchers have explored DRL for various IC scenarios, including non-stationary uncertain demand \citep{dehaybe2024deep, park2023adaptive}, multi-product \citep{sultana2020reinforcement, selukar2022inventory}, variable kinds of products \citep{meisheri2020using,meisheri2022scalable}, multi-echelon supply chains \citep{wu2023distributional, stranieri2024combining}, one-warehouse multi-retailer \citep{kaynov2024deep}, and the stochastic capacitated lot sizing problem \citep{van2023using}. 

However, traditional RL methods suffer from low sample efficiency, presenting a major barrier to real-world implementation where experience collection can be costly and time-consuming \citep{boute2022deep, de2022reward}. Furthermore, this issue is escalated in lost-sales IC problems because of censored demands \citep{chen2024learning}, where customers' real demands are unobservable due to insufficient inventory. For instance, if the order is placed daily, it takes over a year to collect four hundred experiences and part of them may be censored. The scarcity of valid experience presents a true challenge for training data-hungry RL policies. \cite{de2022reward} tries to alleviate this problem by incorporating heuristic knowledge, such as the base-stock policy, into DQN with reward-shaping. Although this method can improve the sample efficiency, it still relies on specific model heuristics, making it hard to generalize to different scenarios. Therefore, overcoming the sample inefficiency of RL methods without assuming strong heuristics remains a key challenge for effectively applying these techniques to IC problems, especially under lost-sales consideration. The detailed related work is provided in Appendix \ref{appendix:related-work}.




To address the above-mentioned limitations of DRL for IC problems, this paper proposes a novel training framework that combines reinforcement learning with feedback graphs (RLFG) and intrinsically motivated exploration (IME):

1) We tailor the feedback graph (FG) based on the general property of lost-sales IC problems rather than case-specific heuristics (e.g. known demand distribution). In particular, the connectivity of FG is adjusted dynamically based on the relationship between the demand and inventory to be adaptive to the lost-sales property. With FG, the sample efficiency in the training process is significantly improved by allowing the RL agent to acquire not only online experiences but also side experiences from FG.

2) We conduct theoretical analysis on how FG reduces the sample complexity with Q-learning to reveal its mechanism of improving the update probabilities across all state-action pairs in IC problems. 

3) Inspired by these theoretical insights, we design a novel intrinsic reward guiding the RL algorithm to explore towards the state-action space where more side experiences can be obtained thereby further boosting sample efficiency. 

4) We evaluate our method across a wide range of IC settings, including the single-item, multi-item, and multi-echelon lost-sales inventory control environments. The empirical results demonstrate the superior sample efficiency and performance improvement of proposed methods, underlining the effectiveness of our designs and generalization ability to different scenarios.

\section{Preliminary}\label{sec:formulation}

\subsection{Lost Sales Inventory Control Problem}

\begin{table}[h!]
    \begin{center}
    \caption{Nomenclature for environmental parameters.}
    \label{table:IC variables}
    \begin{tabular}{l l}
    \hline
     Symbol     & Representation    \\ 
     \hline
     $t=1,...,T$         & Time index    \\
     $a_t\in \mathbb{N}$      & Action: order at time $t$\\  
     $a^{\text{max}}\in \mathbb{N}$  & Maximum order   \\
     $c\in \mathbb{R}$        & Unit cost of procurement     \\
     $d_t\in \mathbb{N}$      & Real demand at time $t$ \\
     $d^o_t\in \mathbb{N}$    & Demand observed at time $t$   \\
     $d^{\text{max}}\in \mathbb{N}$  & Maximum demand  \\
     $h\in \mathbb{R}$  & Unit cost of holding inventory \\
     $L\geq 0$        & Order lead time  \\
     $p\in \mathbb{R}$        & Unit cost penalty of lost sales\\
     $\boldsymbol{s}_t\in \mathbb{N}^{L}$      & State: $(y_t,a_{t+1-L},...,a_{t-1})$ \\
     \multirow{2}{*}{$y_t\in \mathbb{N}$}       & Inventory after receiving  \\
     &the arrived order at time $t$.\\
     \multirow{2}{*}{$y^{\text{max}}\in \mathbb{N}$}  & Maximum inventory after receiving   \\
     & the arrived order at time $t$.
    \\
     $\gamma\in (0,1]$   & Discount factor      \\
    \hline
    \end{tabular}
    \end{center}
\end{table}

We formulate a standard, single-item, discrete-time, and lost-sales IC problem following \cite{zipkin2008old}, \cite{gijsbrechts2022can} and \cite{xin2021understanding} with environment variables defined in Table \ref{table:IC variables}. In lost-sales settings, excess demand is unrecorded when inventory is insufficient, with customers leaving due to unsatisfied demands. The lost-sales IC problem considers three kinds of costs in each time step $t$, which are the cost of procurement $f_1(\boldsymbol{s}_t)$, the cost of holding inventory $f_2(\boldsymbol{s}_t)$, and the cost penalty of lost sales $f_3(\boldsymbol{s}_t)$. The cost terms are specified below,
\begin{center}
\begin{equation}
\begin{aligned}
        &f_1(\boldsymbol{s}_t,a_t) = ca_t,  f_2(\boldsymbol{s}_t)=h[y_t-d_t]^+,\\
        &f_3(\boldsymbol{s}_t)=p[d_t-y_t]^+,
        \text{where }[x]^+=\max(x,0).
        \label{eq:objfun}
\end{aligned}
\end{equation}
\end{center}
In this problem, the real demand is unobservable when it exceeds the current inventory. Thus we define the real demand $d_t$ as a random variable and the observed demand $d_t^o$, which may be censored by $y_t$ and $d_t$, given as

\begin{center}
\begin{equation}
d_t^o=\min(d_t,y_t)= \begin{cases}
        d_t, &\  \text{if}\ d_t\leq y_t \\
        y_t,&\ \text{otherwise}
     \end{cases}.
\label{var:dt}
\end{equation}
\end{center}

Here, the term ``censored'' refers to the case when $d_t>y_t$, i.e. we can only observe all inventory is sold but do not know the value of $d_t$. The objective is to minimize the total cost over the time horizon $T$ under the uncertainty on the demand side, which is
\begin{equation}
\begin{aligned}
    \min_{\{a_t|t=0,...,T\}} &\sum_{t=0}^T\gamma^{t}f(\boldsymbol{s}_t,a_t)\\
    =&\sum_{t=0}^T\gamma^{t}[f_1(\boldsymbol{s}_t,a_t)+f_2(\boldsymbol{s}_t)+f_3(\boldsymbol{s}_t)].
\end{aligned}
\end{equation}

\subsection{MDP Formulation}\label{sec:MDP formulate}

To better understand the lost-sales IC problem from the perspective of RL, we formulate it into an infinite-horizon MDP with discounted rewards \citep{li2020sample}, represented by $\mathcal{M}=(\mathcal{S},\mathcal{A},P,R,\gamma)$.  The MDP consists of the state space $\mathcal{S}$, action space $\mathcal{A}$, transition function $P(\boldsymbol{s}'|\boldsymbol{s},a)$, reward function $R$, and discount factor $\gamma$. The detailed composition of MDP is shown as follows:

\textbf{State:} The state includes the inventory at time $t$ after receiving the orders and all upcoming orders due to lead time. We define it as $\boldsymbol{s}_t=(y_t,a_{t+1-L},...,a_{t-1})$.

\textbf{Action:} The action is the amount to be ordered for future sales, given as $a_t=\{0\leq a_t\leq a^{\text{max}}, a_t\in \mathbb{N}\}$, where $a^{\text{max}}$ is the maximum amount that can be ordered for this item.

\textbf{Reward:} Since the goal is to minimize the cumulative discounted cost, the reward at each time step $t$ is defined as the opposite number of costs, which is $r_t = R(\boldsymbol{s}_t,a_t)=-f(\boldsymbol{s}_t,a_t)$.

\textbf{Transition Function:} The transition function is defined as $\boldsymbol{s}_t=(y_t,a_{t+1-L},...,a_{t-1})\to \boldsymbol{s}_{t+1}=(y_{t+1},a_{t+2-L},...,a_{t})$. The inventory transition is defined as $y_{t+1}=[y_t-d_t]^++a_{t+1-L}$ due to the lead time. As $a_{t+1-L}$ has been received, $a_{t+1+i-L}$ in $\boldsymbol{s}_{t+1}$ will replace $a_{t+i-L}$ in $\boldsymbol{s}_t$ for $i=1,...,L-1$.

\subsection{Reinforcement Learning with Feedback Graph}

RLFG is proposed by \cite{dann2020reinforcement} to reduce the sample complexity of RL. In typical RL scenarios, an agent can only get one experience each time as feedback that can be used to update the policy. However, given prior knowledge about the environments, additional side experiences involving other states and actions may be observable. RLFG explores how RL leverages these side experiences by constructing a feedback graph\footnote{Here ``side experiences'' bears the same meaning as ``side observations'' in \cite{dann2020reinforcement}, and we use the term ``side experiences'' to avoid confusion with ``observations'' in partially observable MDPs.} (FG). FG is denoted as a directed graph $\mathcal{G}=(\mathcal{V}, \mathcal{E})$, where $\mathcal{V}$ is the vertex set $\mathcal{V}=\{\boldsymbol{v}|\boldsymbol{v}=(\boldsymbol{s},a)\}$ and $\mathcal{E}$ is the edge set $\mathcal{E}=\{\boldsymbol{v}\rightarrow \boldsymbol{\hat v}\}$, formalized by the side information indicating that if the agent visits experience $\boldsymbol{v}$, then the corresponding side experiences $\boldsymbol{\hat v}$ are also observable. The total experiences $\mathcal{O}_t$ observed by the agent at timestep $t$ from $\mathcal{G}$ is 

\begin{center}
\begin{equation}
    \mathcal{O}_t(\mathcal{G})=\{(\boldsymbol{s}_t,a_t,r_t,\boldsymbol{s}_{t+1})\}\cup\{\boldsymbol{\hat{s}_t},\hat{a_t},\hat{r_t},\boldsymbol{\hat{s}}_{t+1}\}.
\end{equation}
\end{center}

\section{Method}\label{sec:method}

\begin{figure}[htb]
    \centering 
    \includegraphics[width=0.5\textwidth]{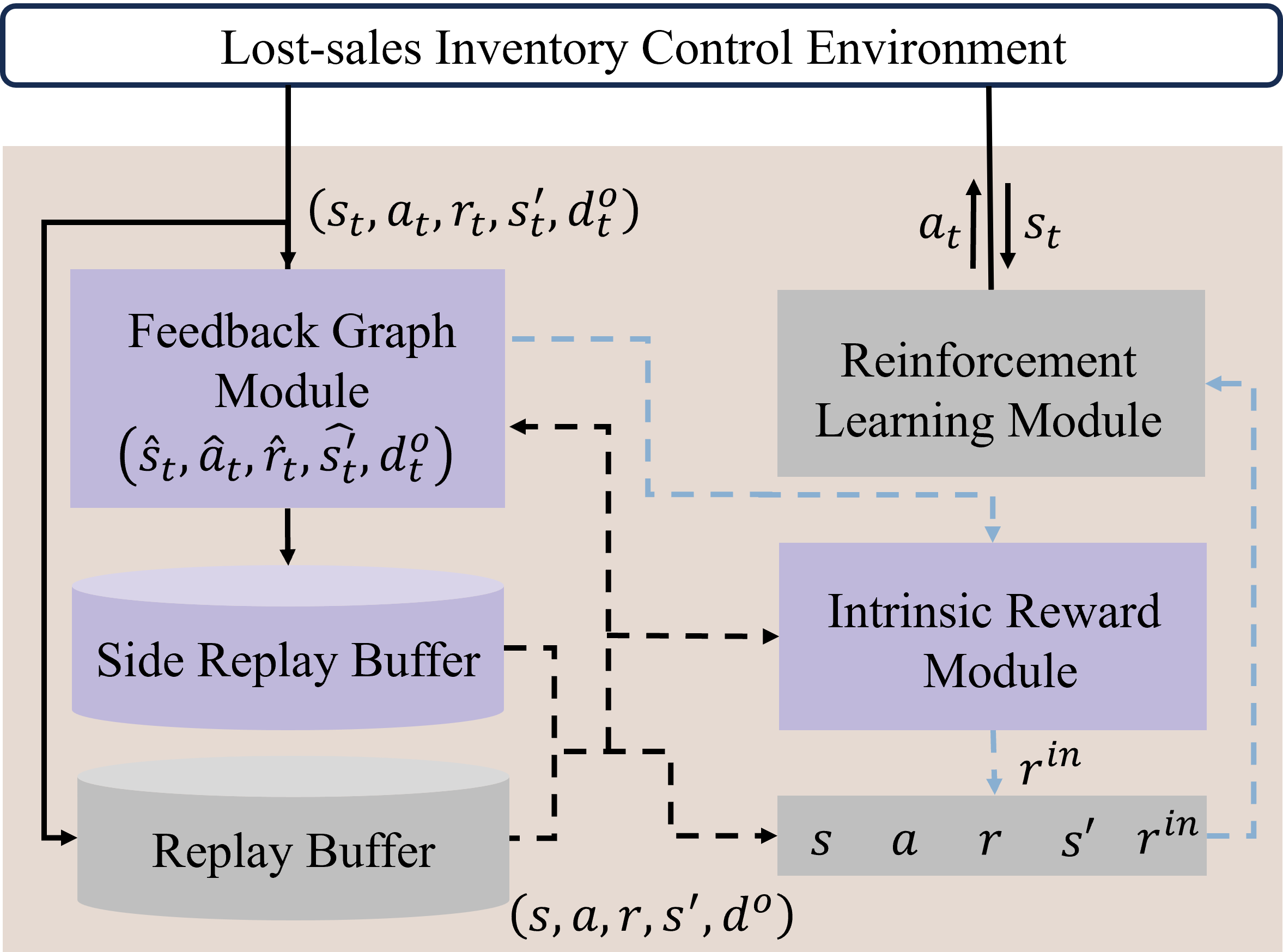} 
    \caption{IC training framework based on RLFG and IME. Solid lines depict sampling steps and dashed lines represent model-updating steps.} 
    \label{fig_sim} 
\end{figure}

\subsection{IC Training Framework with RLFG and IME}
We propose a training framework that incorporates RLFG and IME to enhance the sample efficiency of off-policy RL algorithms in IC problems. This framework relies on two primary assumptions causing sample scarcity. First, the real demand is unknown and may be censored. Second, the experiences collected in real-world operations are limited due to the cost of collecting online data. As illustrated in Figure \ref{fig_sim}, the FG module generates side experiences $\{(\mathbf{\hat{s}_t}, \hat{a_t}, \hat{r_t}, \mathbf{\hat{s}_{t+1}}, d^o_t)\}$ to improve the sample efficiency and the intrinsic reward module aids the exploration by guiding the RL algorithm to explore in state-action space with more side experiences. Here, the RL module can be any off-policy RL algorithm, 
such as DQN, DDPG, Rainbow, or TD3. 
During model-updating steps, a batch of samples $\{(\mathbf{s}, \hat{a}, \hat{r}, \mathbf{s'}, d^o_t)\}$ are sampled from both reply buffers. Subsequently, the FG module generates a new set of side experiences $\{(\mathbf{\hat{s}}, \hat{a}, \hat{r}, \mathbf{\hat{s}'}, d^o_t)\}$, which are exclusively used to calculate the intrinsic rewards of the samples. Rainbow-FG is provided as an example in Algorithm \ref{algo:rainbow-fg} in Appendix \ref{appdendix:algorithm}.

\subsection{Feedback Graph in Inventory Control}\label{sect:FG in IC}

Motivated by the ability of FG to reduce sample complexity \citep{dann2020reinforcement}, we incorporate it into the IC problems. FG is naturally suitable since IC problems follow structured MDP formation and have predictable transitions based on demand, typically independent of state-actions pairs. Observing the demand allows us to leverage the underlying properties of structured MDPs to obtain side experiences. Thus the key question is how to construct FG considering the potentially censored demand due to lost sales. To address this problem, we propose to construct FG dynamically based on the observed demand. 
\begin{center}
\begin{algorithm}[!htb]
	\caption{Feedback Graph Module.} 
	\begin{algorithmic}[1]
	\STATE \textbf{Input:} experience $(\boldsymbol{s}_t=(y_t,a_{t+1-L},...,a_{t-1}),$ $a_t,r_t,\boldsymbol{s}_{t+1},d_t^o)$
        \STATE Initialize a temporary buffer $\mathcal{B}=\{\}$
        \IF {$d_t^o = y_t$ (censored case)}
            \STATE $y^{\text{bound}}=y_t$
        \ELSE
            \STATE $y^{\text{bound}}=y^{\text{max}}$
        \ENDIF
        \FOR{$\hat{y}_t=0,...,y^{\text{bound}}$}
        \FOR{each term of $(\hat{a}_{t+1-L},...,\hat{a}_{t-1})=0,...,a^{\text{max}}$}
        \STATE $\hat{s}_t=(\hat{y}_t,\hat{a}_{t+1-L},...,\hat{a}_{t-1})$
        \FOR{$\hat{a}_t=0,...,a^{\text{max}}$}
        \STATE Get $\hat{r}_t$ and $\hat{\boldsymbol{s}}_{t+1}$ according to $d^o_t$
        \STATE $\mathcal{B}=\mathcal{B}\cup (\hat{\boldsymbol{s}_t},\hat{a_t},\hat{r_t},\hat{\boldsymbol{s}}_{t+1},d_t^o)$
        \ENDFOR
        \ENDFOR
        \ENDFOR
        \STATE Return $\mathcal{B}$
	\end{algorithmic}\label{algo:fgmodule}
\end{algorithm}
\end{center}
Algorithm \ref{algo:fgmodule} outlines the details of the FG module. When $d^o_t$ is uncensored, FG forms a complete graph allowing all state-action pairs to generate side experiences based on $d^o_t$. When $d^o_t$ is censored, only the state-action pairs with smaller inventory levels than $\boldsymbol{s}_t$ can be utilized to generate side experiences, resulting in a partially connected FG. Since $d^o_t$ is censored, the store can only observe that all the inventory are sold out and assume that less inventory can also be sold out without any information for more inventory. Note that FG is still dynamically constructed under the censored case as the number of side experiences depends on $d^o_t$.
\subsection{Theoretical Analysis}\label{sect:theoretical}
We conduct a quantitative analysis of how FG reduces sample complexity in the lost-sales IC environment. Here we provide an analysis based on Q-learning for simplicity. The qualitative analysis based on the property of RLFG is in Appendix \ref{appdendix:analysis-graph-property}.

Without loss of generality, we restrict some definitions in Section \ref{sec:MDP formulate} and define some new notations. We consider the reward function $R:\mathcal{S}\times\mathcal{A}\to(0,1)$. The demand $d_t\sim P_d(d)$ follows any independent discrete distribution and $d^{\text{max}}<y^{\text{max}}$. We define $\pi_b$ as the stationary behavior policy and $\mu(\boldsymbol{s},a)$ as the stationary distribution of the Markov chain under $\pi_b$ and $P(\boldsymbol{s}'|\boldsymbol{s},a)$. In standard RL, $\mu(s,a)$ also indicates the update probability since only visited state-action pairs can be updated. 

\textbf{Scenario 1:} Consider an edgeless graph $\mathcal{G}$ with all state-action pairs as the nodes.

\textbf{Lemma 1:} The sample complexity of the asynchronous Q-learning under scenario 1 is analyzed by \cite{li2020sample}, which is $\Tilde{O}(\frac{1}{\mu_{\text{min}}(1-\gamma)^5\epsilon^2}+\frac{t_\text{mix}}{\mu_{\text{min}}(1-\gamma)})$, where $\mu_{\text{min}}=\min_{(s,a)\in \mathcal{G}}\mu(\boldsymbol{s},a)$ and $t_\text{mix}$ is the mixing time of the Markov chain.

Lemma 1 indicates that the sample complexity of Q-learning without FG is determined by $\mu_{\text{min}}$. In the following parts, we will demonstrate: \textbf{Although the stationary distribution is unchanged, RLFG improves sample efficiency because of increased update probability ($\Tilde{\mu}(\boldsymbol{s},a)\geq\mu(\boldsymbol{s},a)$) due to the side experiences.} From scenario 1 to scenario 6, we generalize the condition from RL to RLFG, which is provided in Appendix \ref{Appendix:Theoretical Analysis}. Here we directly show the final condition of RLFG in scenario 6.

\textbf{Scenario 6:} Consider a graph $\mathcal{G}$ with all state-action pairs as the nodes.  Consider all nodes satisfying $y\geq d_t$, if one is sampled by the policy, all can be sampled by the FG module and updated simultaneously. For nodes with $y<d_t$, only nodes with $y'\leq y$ can be sampled by the FG module and updated simultaneously.

\textbf{Theorem 1:} The update probability for Q-learning with FG under scenario 6 is Equation (\ref{eq:Qlearning-FG-mu}). The proof is provided in Appendix \ref{Appendix:Theoretical Analysis}.

\begin{center}
\begin{equation}
\begin{aligned}
\Tilde\mu(\boldsymbol{s},a)&= \underbrace{\sum_{d=0}^{d^{\text{max}}}P_d(d)\sum_{\substack{(\boldsymbol{\hat s},\hat a)\in \mathcal{G}\\ \hat y\geq d}}\mu(\boldsymbol{\hat s},\hat a|d)}_{\text{Uncensored term}} + \underbrace{\sum_{d=0}^{y-1}P_d(d)\sum_{\substack{(\boldsymbol{\hat s},\hat a)\in \mathcal{G}\\ y\leq\hat y\leq d}}\mu(\boldsymbol{\hat s},\hat a|d)}_{\text{Censored term}}. \label{eq:Qlearning-FG-mu}
\end{aligned}
\end{equation}
\end{center}
\textbf{Remark: We can obtain the relationship between $\Tilde\mu(\boldsymbol{s},a)$ and $\mu(\boldsymbol{s},a)$ in Equation (\ref{Qlearning-FG-mu-compare}), which shows that FG improves the update probability for all the state-action pairs and thus improves the $\mu_\text{min}$, which is the state-action pair with minimum of update probability.}
\begin{equation}
\Tilde\mu(\boldsymbol{s},a)=\mathbb{E}_{d\sim P_d}[\Tilde\mu(\boldsymbol{s},a|d)] \geq \mathbb{E}_{d\sim P_d}[\mu(\boldsymbol{s},a|d)]=\mu(\boldsymbol{s},a). \label{Qlearning-FG-mu-compare}
\end{equation}

\subsection{Intrinsically Motivated Exploration}
To further leverage the benefits of FG, we design a curiosity-driven and FG-driven intrinsic reward to direct the RL algorithm towards the state-action space where more side experiences can be generated. For each state-action pair, the quantity and average uncertainty of the side experiences generated by this state-action pair are incorporated into the intrinsic reward.

This idea is motivated by the theoretical analysis based on Equation \ref{eq:Qlearning-FG-mu}. The uncensored term reflects the contribution of the uncensored case to improve the update probability. Enhancing this term can improve the update probability of all state-action pairs. The censored term measures the benefits a state-action pair gains from the censored cases. In this scenario, the update probability can be improved from the state-action pairs with larger inventory levels. To satisfy both conditions, manually designing a behavior policy is difficult and not general enough. Instead, we propose an intrinsic reward given in Equation \ref{eq:inre} promoting side experience generation.
\begin{center}
\begin{equation}
r^{in}_i=r^{in}_i+log_{10}(J)\times\frac{1}{J}\sum_{j=1}^Jr^{in}_{i,j}, \label{eq:inre}
\end{equation}
\end{center}
where $J$ is the number of the side experiences generated by an experience.
Algorithm \ref{algo:inr} shows the details of the intrinsic reward module. We utilize the M-head DQN \citep{nikolov2018information} with each head trained by different mini-batches of experiences to get the prediction error as the curiosity. The final intrinsic reward, denoted in Equation \ref{eq:inre}, consists of the curiosity of the experience itself ($r^{in}_i$) and the averaged curiosity of all corresponding side experiences ($r^{in}_{i,j}$). The averaged value preserves the advantage of scale invariance but loses the quantity information compared with the sum value. To balance these two aspects, we multiply $\log_{10}(J)$ with the averaged value to incorporate more quantity information into the intrinsic reward. In the censored case, where $J=y_t$, a larger $J$ increases the intrinsic reward, which means more side experiences. In the uncensored case, $J$ has no differences, and a larger intrinsic reward indicates higher curiosity. Across the censored and uncensored case, $J=y^{\text{max}}$ is larger than $y_t$, which means that the intrinsic reward in the uncensored case is more likely to be larger than that in the censored case.

\begin{center}
\begin{algorithm}[h!]
	\caption{Intrinsic Reward Module.} 
	\begin{algorithmic}[1]
	\STATE \textbf{Input:} Mini-batch experiences $\{(\boldsymbol{s},a,r,\boldsymbol{s}',d^o)\}$ and all side experiences $\{(\hat{\boldsymbol{s}},\hat{a},\hat{r},\hat {\boldsymbol{s}}',d^o)\}$
        \FOR{each experience $(\boldsymbol{s}_i,a_i,r_i,\boldsymbol{s}'_i,d^o_i)$ in $\{(\boldsymbol{s},a,r,\boldsymbol{s}',d^o)\}$}
        \STATE $\Bar{Q}^{dqn}=\frac{1}{M}\sum_{m=1}^MQ^{dqn}_m(\boldsymbol{s}_i,a_i)$ 
        \STATE $r^{in}_i=\frac{1}{M}\sqrt{\sum_{m=1}^M[Q^{dqn}_m(\boldsymbol{s}_i,a_i)-\Bar{Q}^{dqn}]^2}$
        \STATE $\{(\hat{\boldsymbol{s}}_i,\hat{a}_i,\hat{r}_i,\hat {\boldsymbol{s}}'_i,d^o_i)\}$ are all side experiences generated from $(\boldsymbol{s}_i,a_i,r_i,\boldsymbol{s}'_i,d^o_i)$
        \STATE $J$ is the size of $\{(\hat{\boldsymbol{s}}_i,\hat{a}_i,\hat{r}_i,\hat {\boldsymbol{s}}'_i,d^o_i)\}$
        \FOR{each side experience $(\hat{\boldsymbol{s}}_{i,j},\hat{a}_{i,j},\hat{r}_{i,j},\hat{\boldsymbol{s}}'_{i,j},d^o_i)$ in $\{(\hat{\boldsymbol{s}}_i,\hat{a}_i,\hat{r}_i,\hat {\boldsymbol{s}}'_i,d^o_i)\}$}
        \STATE $\Bar{Q}^{dqn}=\frac{1}{M}\sum_{m=1}^MQ^{dqn}_m(\hat{\boldsymbol{s}}_{i,j},\hat{a}_{i,j})$ 
        \STATE $r^{in}_{i,j}=\frac{1}{M}\sqrt{\sum_{m=1}^M[Q^{dqn}_m(\hat{\boldsymbol{s}}_{i,j},\hat{a}_{i,j})-\Bar{Q}^{dqn}]^2}$
        \ENDFOR
        \STATE $r^{in}_i=r^{in}_i+log_{10}(J)\times\frac{1}{J}\sum_{j=1}^Jr^{in}_{i,j}$
        \STATE Store $r^{in}_i$ into $(\boldsymbol{s}_i,a_i,r_i,\boldsymbol{s}'_i,d^o_i)$
        \ENDFOR
        \STATE Return $\{(\boldsymbol{s},a,r,\boldsymbol{s}',d^o,r^{in})\}$
	\end{algorithmic}\label{algo:inr}
\end{algorithm}
\end{center}

\begin{table*}[htb] 
    \begin{center}
    \caption{Optimal result comparison on single-item environment. For each method, the first row is the optimal average cost ($\downarrow$) and the second row is the optimality gap ($\downarrow$). The bold values indicates the values of the best five methods. $p$ denotes the lost-sales penalty, and $L$ denotes the lead time.}
    \label{table:opt-re}
    \setlength{\tabcolsep}{3mm}{
    \begin{tabular}{ c c c c c c c }
    \hline
     \multirow{2}{*}{Method}      & \multicolumn{3}{c|}{$p$=4}      & \multicolumn{3}{c}{$p$=9}  \\ 
     \cline{2-7}
     &\multicolumn{1}{c|}{$L$=2} & \multicolumn{1}{c|}{$L$=3} & \multicolumn{1}{c|}{$L$=4} &\multicolumn{1}{c|}{$L$=2} & \multicolumn{1}{c|}{$L$=3} & \multicolumn{1}{c}{$L$=4}\\ 
     \hline
     Optimal     & 4.40        & 4.60        &4.73        &6.09   &6.53               
     &6.84     \\
     \hline
     \multirow{2}{*}{Constant Order}     & 5.27 & 5.27 &5.27 &10.27 &10.27 &10.27    \\
                        & 19.8\% & 14.6\% &11.4\% &68.6\% &57.3\% &50.1\%    \\
    \hline
     \multirow{2}{*}{Bracket}            &5.00  &5.01 &5.02 &8.02 &8.02 &8.03    \\
                        &13.6\% &8.9\% &6.1\% &31.7\% &22.8\% &17.3\%    \\
    \hline
     \multirow{2}{*}{Myopic 1-period}     & 4.56 & 4.84 &5.06 &6.22 &6.80 &7.20    \\
                         & 3.7\% & 5.3\% &7.1\% &2.1\% &4.1\% &5.3\%    \\
    \hline
     \multirow{2}{*}{Myopic 2-period}     & \textbf{4.41} & \textbf{4.64} &\textbf{4.82} &\textbf{6.10} &\textbf{6.57} &\textbf{6.92}    \\
          & \textbf{0.2\%} & \textbf{0.8\%} &\textbf{1.9\%} &\textbf{0.2\%} &\textbf{0.6\%} &\textbf{1.2\%}    \\
     \hline
     \multirow{2}{*}{Base-Stock}     & 4.64 & 4.98 &5.20 &6.32 &6.86 &7.27    \\
                    & 5.5\% & 8.2\% &9.9\% &3.7\% &5.1\% &6.4\%    \\
     \hline
     \multirow{2}{*}{Capped Base-Stock} & \textbf{4.41} & \textbf{4.63} &\textbf{4.80} &\textbf{6.12} &\textbf{6.62} &\textbf{6.91}    \\
        & \textbf{0.2\%} & \textbf{0.7\%} &\textbf{1.5\%} &\textbf{0.5\%} &\textbf{1.4\%} &\textbf{1.0\%}    \\
     \hline\hline
     \multirow{2}{*}{A3C}     & 4.54 & 4.74 &5.05 &6.38 &6.73 &7.07    \\
             & 3.2\% & 3.0\% &6.7\% &4.8\% &3.1\% &3.4\%    \\
     \hline
     \multirow{2}{*}{Rainbow}     & 4.52 & 4.72 &4.91 &6.28 &6.72 &7.09    \\
                 & 2.7\% & 2.6\% &3.8\% &3.1\% &2.9\% &3.7\%    \\
     \hline
     \multirow{2}{*}{Rainbow-FG (ours)}  & \textbf{4.48} & \textbf{4.67} &\textbf{4.87} &\textbf{6.22} &\textbf{6.73} &\textbf{6.99}    \\
                 & \textbf{1.8\%} & \textbf{1.5\%} &\textbf{2.9\%} &\textbf{2.1\%} &\textbf{3.1\%} &\textbf{2.2\%}    \\
     \hline
     \multirow{2}{*}{TD3}     & 4.53 & 4.76 &4.87 &6.29 &6.78 &7.07    \\
                 & 2.9\% & 3.4\% &2.9\% &3.2\% &3.8\% &3.3\%    \\
     \hline
     \multirow{2}{*}{TD3-FG (ours)}  & \textbf{4.42} & \textbf{4.62} &\textbf{4.76} &\textbf{6.14} &\textbf{6.62} &\textbf{6.90}    \\
                 & \textbf{0.4\%} & \textbf{0.4\%} &\textbf{0.6\%} &\textbf{0.8\%} &\textbf{1.4\%} &\textbf{0.9\%}    \\
     \hline
     \multirow{2}{*}{Rainbow-FG (H) (ours)} & \textbf{4.404} & \textbf{4.61} &\textbf{4.775} &\textbf{6.12} &\textbf{6.60} &\textbf{6.92}    \\
                    & \textbf{0.1\%} & \textbf{0.2\%} &\textbf{0.9\%} &\textbf{0.5\%} &\textbf{1.1\%} &\textbf{1.2\%}    \\
    \hline
    \end{tabular}
    }
    \end{center}
\end{table*}

\section{Experiment}\label{sec:experiment}
To verify the performance and sample efficiency of our training framework, we test our method, including (1) \textbf{Rainbow-FG}, (2) \textbf{TD3-FG}, and (3) \textbf{Rainbow-FG (H)}, on the (a) \textbf{single-item}, (b) \textbf{multi-item}, and (c) \textbf{multi-echelon} lost-sales IC environment. Rainbow-FG (H) is a variant, which utilizes the heuristic knowledge to continue finetuning Rainbow-FG. It can be regarded as the upper bound of Rainbow-FG.  All experiments are averaged on 20 random seeds with shaded areas representing the standard deviation. 

\subsection{Baselines}
\textbf{Heuristic Methods:} We compare our method with widely recognized and well-performed heuristic methods with optimized parameters via brute force search. 

\begin{itemize}[leftmargin=0.3cm, itemindent=0cm]
\item \textbf{Constant Order}: Fixed order $a_t=r^h$ with $r^h$ being a parameter.
\item \textbf{Myopic 1-period} (\cite{morton1971near}): Assume the distribution of $d_t$ is known. This method aims to minimize the expected cost at $t+L$. The order is $a_t=\min_{P(y_{t+L}-d_{t+L}<0)\leq \frac{c+h}{p+h}} a_t$, where $0\leq a_t\leq a^{\text{max}}$.
\item \textbf{Myopic 2-period} (\cite{zipkin2008old}): This method considers 2-step expected costs from $t+L$ compared with Myopic 1-period method.
\item \textbf{Base-Stock method} (\cite{zipkin2008old}): This method aims to maintain the inventory level with considering upcoming orders. The order is $a_t=(S^h-\bm{1}\cdot \boldsymbol{s}_t)^+$, where $S^h$ is a parameter.
\item \textbf{Capped Base-Stock method} (\cite{xin2021understanding}): It combines Base-Stock and Constant Order method. The order is $a_t=\min[(S^h-\boldsymbol{1}\cdot\boldsymbol{s}_t)^+, r^h]$, where $S^h$ and $r^h$ are parameters.
\item \textbf{Bracket method} (\cite{bai2023asymptotic}): A variant of the Constant Order method. The order is $a_t = \lfloor(t+1)r^h+\theta^h\rfloor-\lceil tr^h+\theta^h\rceil$, where $r^h$ and $\theta^h$ are parameters.
\end{itemize}

\textbf{Deep RL Methods:}
We choose to compare (1) Rainbow; (2) TD3; (3) the A3C method in \cite{gijsbrechts2022can}, which is an on-policy RL method. Since A3C is on-policy, its sample efficiency cannot be directly compared to the off-policy methods. Thus A3C only serves for comparing optimal performance.

    

\begin{figure*}[t]
\begin{minipage}{1.03\linewidth}
    \centering
    \hspace{-6mm}
        \includegraphics[width=1\textwidth]{ 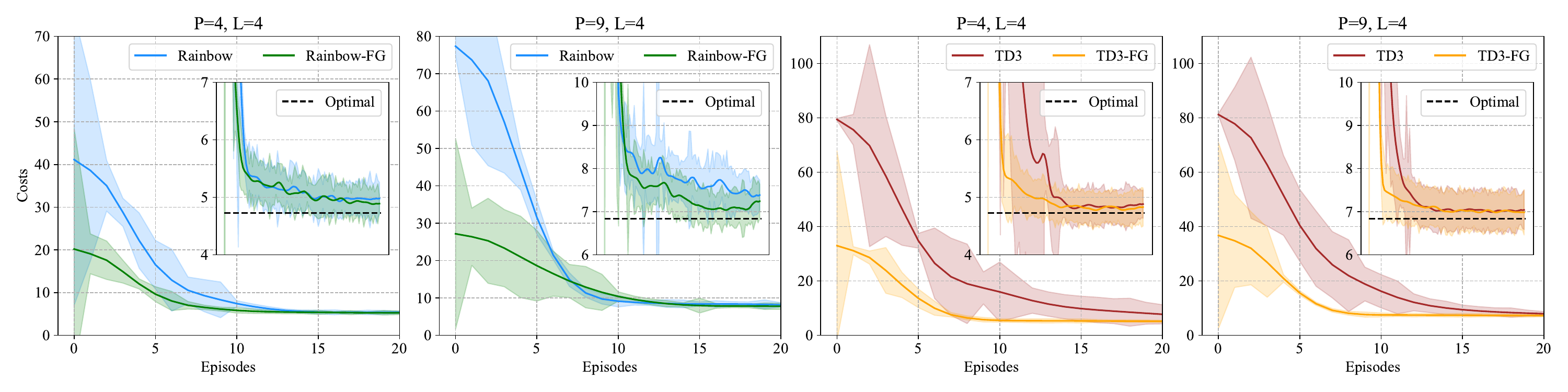}
        \caption{Single-item.}
        \label{exp-single-item}

\end{minipage}
\end{figure*}

\begin{figure*}[t]
    \begin{minipage}{1\linewidth}
        \centering
            \includegraphics[width=1\textwidth]{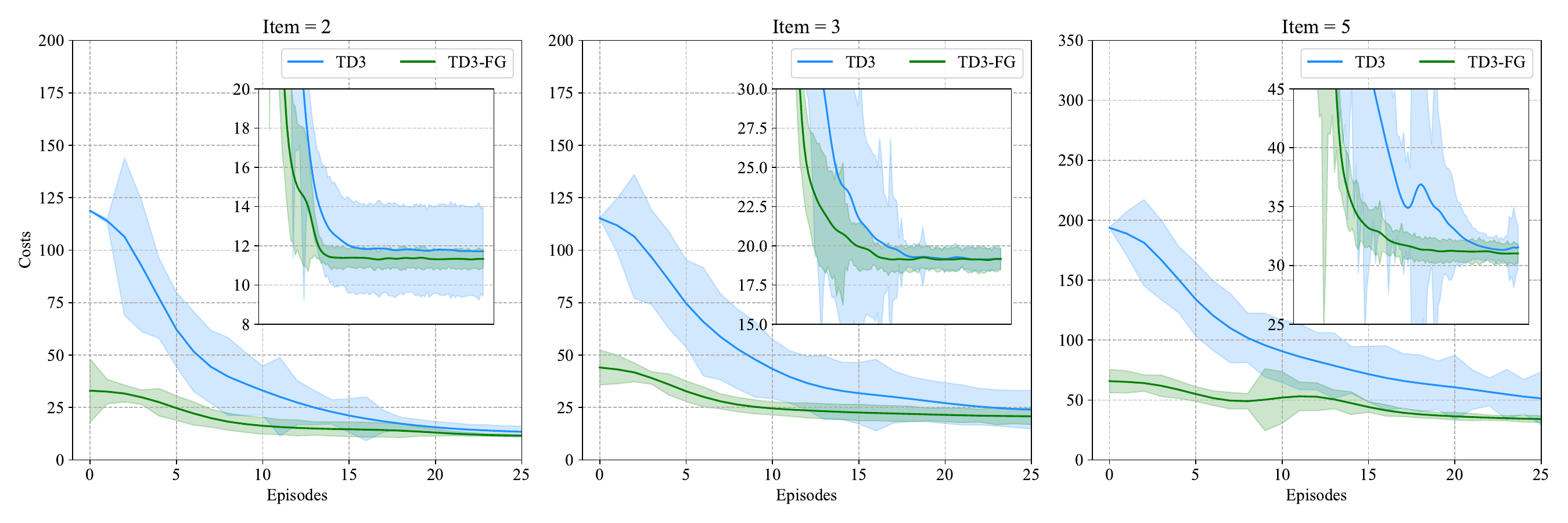}
            \caption{Multi-item.}
            \label{exp-multi-item}

            \includegraphics[width=1\textwidth]{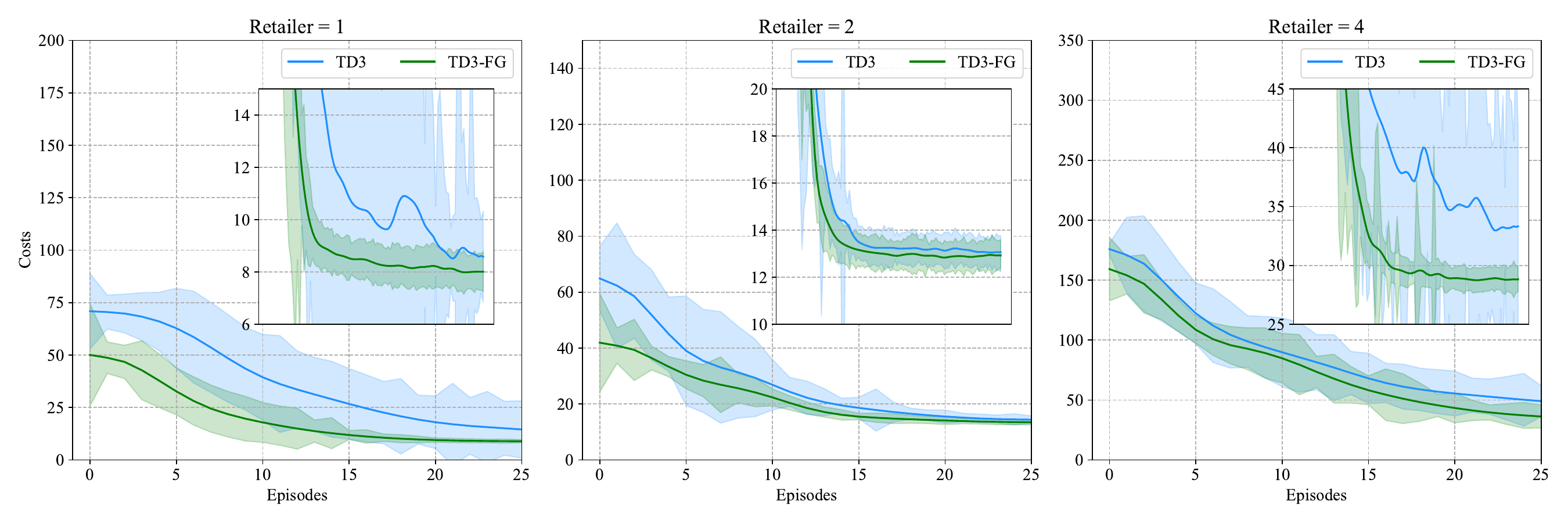}
            \caption{Multi-echelon.}
            \label{exp-multi-echelon}

    \end{minipage}
    \label{exp-fig}
\end{figure*}

\subsection{Case One: Single Item}
We test our methods and baselines across different settings of the single-item lost-sales IC environment with demand obeying Poisson distribution. The parameters of the environment and the hyperparameters of our methods are in Appendix \ref{appendix:para-of-rainbow-fg}. In the following sections, default parameters are used unless stated otherwise. The hyperparameter analysis is demonstrated in Appendix \ref{appendix:hyper-analy}.

\subsubsection{Optimal Results Comparison}
Table \ref{table:opt-re} presents the average results and optimality gap compared to the optimal results. Without strong heuristics, Rainbow, TD3 and A3C demonstrates similar performance whereas Rainbow-FG and TD3-FG achieves superior results. In particular, TD3-FG outperforms Myopic 2-period and Capped Base-Stock method and achieve SOTA performance in some settings without any strong heuristics. This notable improvement can be attributed to the ability to leverage a broader range of side experiences from FG, enabling TD3-FG to converge towards better solutions compared with TD3, and even existing best methods.

Although Rainbow-FG does not surpass the top-performing heuristic methods like Myopic 2-period and Capped Base-Stock, Myopic 2-period method assumes the distribution of the demand is known, which is a strong assumption, and the Capped Base-Stock method needs extensive parameter searches to attain this optimal result. These parameters vary across different scenarios without patterns. Conversely, Rainbow-FG consistently achieves near-optimal performance and possesses adaptability to diverse settings without these strong heuristics. Furthermore, when equipped with the heuristic knowledge, Rainbow-FG (H) attains equivalent or even superior performance compared to the best heuristic methods, particularly in settings ($p=4,L=2,3,4$). This result indicates that Rainbow-FG has the potential to perform better than heuristic methods.


\subsubsection{Sample Efficiency Comparison}\label{subsec:exp-fg-single-item}
Figure \ref{exp-single-item} illustrates the learning process of Rainbow versus Rainbow-FG and TD3 versus TD3-FG under two settings. The utilization of FG significantly enhances the sample efficiency. In the early stages of training, Rainbow-FG and TD3-FG show great sample efficiency compared with Rainbow and TD3.  
Towards the end, Rainbow-FG and TD3-FG converge faster and better compared to Rainbow and TD3, where Rainbow and TD3 fail to converge to the level of their FG version throughout the entire 100 episodes in some settings. Furthermore, FG contributes to improving the stability of the learning process. Notably, Rainbow-FG and TD3-FG exhibit lower standard deviation and more stable learning curves for the whole learning process. Experiments on more settings are shown in Appendix \ref{appendix:more-exp-fg}.
\subsection{Case Two: Multi Item}
We further conduct experiments with TD3 and TD3-FG on the more challenging multi-item lost-sales IC problem with 2, 3, and 5 items in one store. Note that Rainbow and Rainbow-FG is not included in the comparison due to its exponential output growth with items. The parameters of the environment are in Appendix \ref{appendix:para-of-rainbow-fg}.

\subsubsection{Optimal Results Comparison}
Figure \ref{exp-multi-item-echelon-final} shows the final results in the multi-item environment. It demonstrates that TD3-FG can achieve better performance than TD3 in all cases. In particular, TD3-FG decreases the performance gap by 4.58\%, 1.16\%, and 2.56\% for environments with 2, 3, and 5 items. 

\subsubsection{Sample Efficiency Comparison}\label{subsec:exp-fg-multi-item}
Figure \ref{exp-multi-item} demonstrates similar learning processes of TD3 and TD3-FG with different numbers of items like sing-item environment. TD3-FG shows great sample efficiency and better stability throughout the training. As items increase, the efficiency gap between TD3 and TD3-FG widens, while TD3-FG maintains consistent gains, highlighting the scalability of the proposed methods.

\subsection{Case Three: Multi Echelon}
We compare TD3 and TD3-FG on multi-echelon lost-sales IC with 1 warehouse and 1, 2 or 4 retailers to evaluate generalization. The retailers order from the warehouse and satisfy the demand of customers, and the warehouse orders from the manufacturer and satisfies the demand of retailers. Both the warehouse and retailers need to find optimal decisions. The parameters of the environment are in Appendix \ref{appendix:para-of-rainbow-fg}.



\subsubsection{Optimal Results Comparison}
\begin{figure}[h!]
    \centering
    \begin{minipage}{0.5\linewidth}
        \centering
            \includegraphics[width=0.7\textwidth]{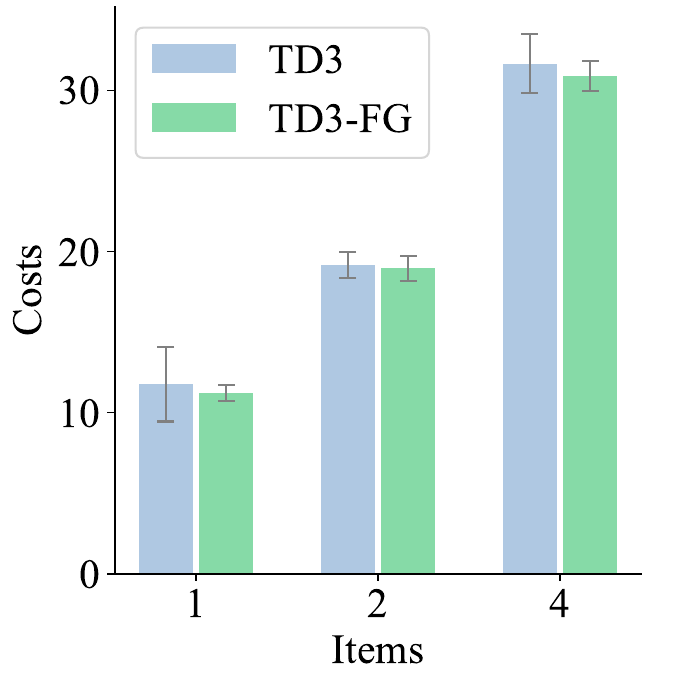}
    \end{minipage}
    \hspace{-5mm}
    \begin{minipage}{0.5\linewidth}
        \centering

            \includegraphics[width=0.7\textwidth]{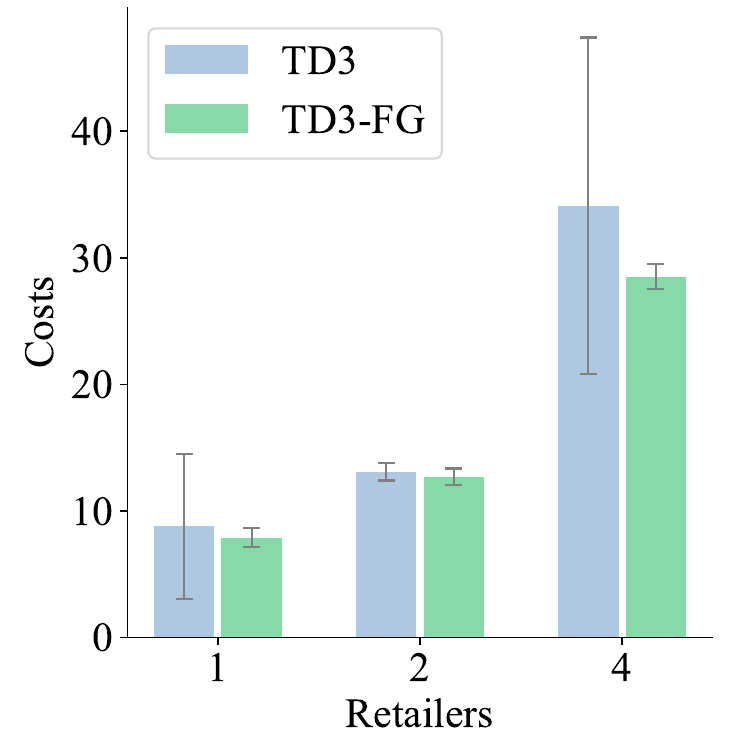}

    \end{minipage}
    \caption{Final results on multi-item and multi-echelon environments.}
    \label{exp-multi-item-echelon-final}
\end{figure}

Figure \ref{exp-multi-item-echelon-final} shows the final results in the multi-echelon environment. It demonstrates that TD3-FG can achieve lower costs than TD3 with lower variance. In particular, TD3 performs very poorly when the number of retailers is 4. This results from failed warehouse decisions, as its demand comprises retailer orders. Since the demand for the warehouse is the sum of the orders from retailers, the mean and variance of the demand for the warehouse also increase as the number of retailers increases. Due to failed warehouse decisions, the retailers also cannot find optimal solutions to satisfy the demand of customers. Thus both the warehouse and retailers are in a greatly lost-sales condition.
\subsubsection{Sample Efficiency Comparison}\label{subsec:exp-fg-multi-echelon}
Figure \ref{exp-multi-echelon} demonstrates similar learning curves of TD3 and TD3-FG with different numbers of retailers like multi-item environment. TD3-FG always achieves better performance with smaller variance under the same training budgets. Besides, TD3-FG converges to near-optimal solutions within around 20 episodes but TD3 converges to worse solutions with 80 to 100 episodes when the numbers of retailers are 1 and 4.  

\section{Conclusion}\label{sec:conclusion}

This paper addresses the challenge of sample scarcity for RL methods in lost-sales IC problems. We design a novel training framework integrating RLFG and IME to boost the sample efficiency of RL methods. We first specially tailored FG using general lost-sales properties, environment-specific heuristics, to generate side experiences to aid RL updates. We then conduct a theoretical analysis to demonstrate our method's effectiveness with Q-learning as an example. The analysis shows that FG decreases the sample complexity by improving the update probabilities for all state-action pairs. Additionally, we design an intrinsic reward to fully utilize FG for lost-sales IC problems based on the theoretical insights. To validate the performance and generalization ability of our proposed method, we have done experiments on single-item, multi-item, and multi-echelon lost-sales IC environments. Experimental results show that our approach greatly enhances RL's sampling efficiency in IC problems, which is consistent with the theoretical analysis.

\bibliographystyle{abbrvnat}
\bibliography{paper}

\newpage

\begin{APPENDICES}
\crefalias{section}{appendix}
\section{Related Work}\label{appendix:related-work}
\subsection{Inventory Control Problem}
The inventory control (IC) problem involves determining the optimal quantity of inventory to order to minimize costs while maintaining sufficient stock levels to meet customer demand. Based on different assumptions about the behavior of the customer, the IC problem can be divided into backlogging IC and lost-sales IC problems \citep{chen2024learning}. The backlogging IC problem assumes that when the demand cannot be met due to insufficient inventory, the request of customers is accepted but delayed until inventory is replenished. \cite{arrow1958studies} has proven that the base-stock policy, which aims to keep the sum of inventory level and upcoming orders constant, is optimal for single-source backlogging IC with constant lead time. Compared with backlogging IC, the lost-sales IC problem is more complex and relevant \citep{bijvank2011lost}. the lost-sale IC problem assumes that when demand cannot be met due to insufficient inventory, the customer's request is lost and the exceeding request is unobservable, such as e-commerce. The base-stock policy can only be optimal when the cost of the lost-sales penalty is high. This motivates researchers to find better methods for the lost-sales IC problem. 

\subsection{Heuristic Methods in Lost-sales Inventory Control}
The lost-sales IC problem is first simply studied by \cite{karlin1958inventory}, which assumes the lead time to place orders is one. \cite{karlin1958inventory} proves that base-stock policy is not optimal because the inventory availability in future periods cannot be characterized by the inventory level and order quantity. \cite{morton1969bounds} extends the former analysis to any positive and integral-value lead time and provides the upper and lower bounds of the optimal policy. Furthermore, \cite{zipkin2008structure} generalizes to any lead times from the aspect of L-natural-convexity, and \cite{janakiraman2004lost} studies the case when lead time is stochastic.  
The base-stock policy is sensitive to the demand due to its design. The opposite one is the Constant Order policy, whose decision has no relationship with the demand \citep{bijvank2023lost}. As the lead time goes to infinity, \cite{goldberg2016asymptotic} proves that the constant-order policy can be asymptotically optimal and \cite{reiman2004new} proves that the constant-order policy can be better than the base-stock policy. The contrary properties of these two methods motivate a better idea that combines both advantages \citep{bijvank2012periodic}. \cite{xin2021understanding} improves this idea to a new method named capped base-stock policy. 

The above-introduced methods need to search for the best parameter for specific settings to perform well. Besides them, there is another series of heuristic method, also called approximate dynamic programming (APD). These methods assume known demand distribution rather than parameter searching. \cite{morton1971near} proposes the Myopic method, which aims to minimize the expected cost when the placed order arrives. This method can be extended to the Myopic-T method, which considers the expected cost for $T$ time steps. 

\subsection{Reinforcement Learning Method in Inventory Control}
The heuristic method needs either a parameter search or an assumption about the demand distribution, which are not general enough. This motivates the studies about applying reinforcement learning (RL) to lost-sales IC problems. RL has gained significant attention as a powerful data-driven technique for solving complex sequential decision-making problems \citep{han2023deep, jiang2024achieving, xie2024vc}. RL offers several advantages for addressing the challenges of IC problems. Firstly, RL allows for the discovery of optimal policies without relying on strong problem-specific assumptions, enabling more generalizable solutions \citep{nian2020review}. Secondly, combined with the deep neural network, Deep RL (DRL) can handle large state and action spaces, making it suitable for problems with high-dimensional variables \citep{dehaybe2024deep}. Initially, the work mainly focuses on verifying the feasibility of RL in IC problems. \cite{oroojlooyjadid2022deep} showcases the ability of a Deep Q-Network (DQN) to discover solutions that are close to optimal for the widely recognized beer distribution game. \cite{gijsbrechts2022can} introduces the A3C algorithm into IC problems and show that A3C can achieve acceptable performance but not better heuristic methods. \cite{stranieri2023comparing} benchmarks various DRL methods such as A3C, PPO, and vanilla policy gradient (VPG) in IC problems. Recently, researchers tend to study how DRL can solve different situations of IC problems such as, non-stationary uncertain demand \citep{dehaybe2024deep, park2023adaptive}, multi-product \citep{sultana2020reinforcement, selukar2022inventory}, variable kinds of products \citep{meisheri2020using, meisheri2022scalable}, multi-echelon supply chains \citep{wu2023distributional, stranieri2024combining}, one-warehouse multi-retailer \citep{kaynov2024deep}, stochastic capacitated lot sizing problem \citep{van2023using}.

However, Two fundamental issues have not been resolved. First, the final performance of DRL is not optimal, sometimes even worse than heuristic methods. To address this problem, some papers \citep{cuartas2023hybrid, stranieri2024combining, harsha2021math} explore the potential to combine RL algorithms with existing other methods rather than directly applying RL to IC problems. The other problem is the low sample efficiency nature of existing RL methods, which restricts further application to real-world IC problems \citep{boute2022deep, de2022reward}, especially when obtaining experiences is expensive or time-consuming. Besides, this sample inefficiency problem is enlarged in lost-sales IC because of censored demands \citep{chen2024learning}, which refers to the phenomenon when the customer's real demand is unobservable due to insufficient inventory. Typically, if the order is placed daily, generating four hundred experiences needs over a year and part of them may be censored making them too hard to update RL policy. \cite{de2022reward} tries to alleviate this problem by incorporating heuristic knowledge, such as base-stock policy, into DQN with reward-shaping. Although this method can improve the sample efficiency, it still relies on specific model heuristics, making it hard to generalize to different scenarios. Overall, Resolving the low sample efficiency of RL without strong heuristics is crucial in solving IC (especially lost-sales IC) problems.

\subsection{Feedback Graph and its Application}
\cite{mannor2011bandits} first proposes the feedback graph (FG) idea as a method to reduce the regret bound of the bandit problem when side observations can be obtained assuming the decision maker can also know the situations when other actions are taken besides the chosen action. \cite{alon2015online} extends the analysis of FG on bandit problems beyond the learning problems. \cite{tossou2017thompson} gives the first analysis of Thompson sampling for bandits with FG based on information theory. Building upon these foundations, \cite{dann2020reinforcement} combines FG with RL and shows how FG can reduce the regret bound and sample complexity of model-based RL algorithms. However, most existing work, such as \cite{liu2018information}, \cite{hao2022contextual}, and \cite{marinov2022stochastic}, mainly focus on the analysis of FG in some bandits problems. Seldom work tries to apply FG to real-world problems since side observations and the structure of FG are hard to define in applications.

\section{Rainbow-FG Algorithm}\label{appdendix:algorithm}
Algorithm \ref{algo:rainbow-fg} shows the example of utilizing FG, called Rainbow-FG.
\begin{center}
\begin{algorithm}[!htb]
	\caption{Rainbow-FG Algorithm.} 
	\begin{algorithmic}[1]
	\STATE Initialize the distributional Q-network $Q$ with parameter $\theta$, the target Q-network $Q'$ with parameter $\phi$, the replay buffer $\mathcal{R}$, and the side replay buffer $\mathcal{R}^s$
	\FOR {episode = $1,...,M$}
        
        \STATE Initialize the environment state $s_0$
        \FOR {$t=0,...,T$}
		  \STATE Get the order $a_t=
                \begin{cases}
                    \text{argmax}_a\ Q(\boldsymbol{s}_t,a),&\  w.p\ \ 1-\epsilon \\
                    \text{a random action},&\ w.p\ \ \epsilon
                \end{cases}$
            \STATE Execute action $a_t$ and get next state $\boldsymbol{s}_{t+1}$ and reward $r_t$
            \STATE Get the observed demand $d_t^o$
            \STATE Get the side experiences $\{(\hat{\boldsymbol{s}_t},\hat{a_t},\hat{r_t},\hat{s}_{t+1},d^o_t)\}\to$ Algorithm \ref{algo:fgmodule} 
            \STATE $\mathcal{R}=\mathcal{R}\cup(\boldsymbol{s}_t,a_t,r_t,\boldsymbol{s}_{t+1},d_t^o)$ and $\mathcal{R}^s=\mathcal{R}^s\cup \{(\hat{\boldsymbol{s}_t},\hat{a_t},\hat{r_t},\hat{\boldsymbol{s}}_{t+1},d_t^o)\}$
            \\
            \STATE \textbf{Model Update:}
            \STATE Randomly sample a batch of experiences $\{(\boldsymbol{s},a,r,\boldsymbol{s}',d^o)\}$ from $\mathcal{R} \cup \mathcal{R}^s$
            \STATE Get all side experiences $\{(\hat{\boldsymbol{s}},\hat{a},\hat{r},\hat {\boldsymbol{s}}',d^o)\}$ for each experience in $\{(\boldsymbol{s},a,r,\boldsymbol{s}',d^o)\}\to$ Algorithm \ref{algo:fgmodule}
            \STATE Get the intrinsic reward $r^{in}$ of this batch of experiences $\to$ Algorithm \ref{algo:inr}
            \STATE Get the final reward $r=(1-\beta)\times r+\beta\times r^{in}$
            \STATE Update parameters
        \ENDFOR
	\ENDFOR
	\end{algorithmic}\label{algo:rainbow-fg}
\end{algorithm}
\end{center}

\section{Analysis based on the Property of Graph}\label{appdendix:analysis-graph-property}
Based on the feedback graph $\mathcal{G}$, \cite{dann2020reinforcement} defines three concepts to measure the sample complexity. $\omega$, $\alpha$, and $\zeta$ measure the minimum vertices to be sampled to observe the whole $\mathcal{G}$. We will talk about $\omega$, $\alpha$, and $\zeta$ for the IC environment in section \ref{sect:FG in IC}.

\textbf{Mas-number($\omega$):} The maximum size of $\mathcal{V}'\subseteq \mathcal{V}$ forming an acyclic subgraph of $\mathcal{G}$ is the mas-number.

\textbf{Independence number($\alpha$):} The size of the largest $\mathcal{V}'\subseteq \mathcal{V}$ with no edge in $\mathcal{G}$ within $\mathcal{V}'$ is the independence number.

\textbf{Domination number($\zeta$):} A set of vertices $\mathcal{V}' \subseteq \mathcal{V}$ is a dominating set if there always $\exists v' \in \mathcal{V}'$ and $\forall v \in \mathcal{V}$ such that $v' \rightarrow v$. The smallest size of $\mathcal{V}'$ is called the domination number $\zeta$.  

For any feedback graph $\mathcal{G}$, the inequality (\ref{eq:graph-ineq}) is always satisfied, where $|\mathcal{S}|$ is the size of the state set and $|\mathcal{A}|$ is the size of the action set. The mas-number $\omega$ and independence number $\alpha$ quantify the worst-case connectivity by measuring the maximum number of distinct vertices an algorithm can traverse before encountering a repeated vertex. On the other hand, the domination number represents a best-case scenario and indicates the minimum number of vertices that an algorithm must visit to observe every vertex.  Besides, \cite{dann2020reinforcement} also shows that with feedback graph $\mathcal{G}$, the regret bound/sample complexity of a model-based RL algorithm can be reduced from $|\mathcal{S}||\mathcal{A}|$ scale to $\omega$ or even $\zeta$ scale. 

\begin{equation}
    \zeta\leq\alpha\leq\omega\leq|\mathcal{V}|=|\mathcal{S}|\times|\mathcal{A}|. \label{eq:graph-ineq}
\end{equation}

In this paragraph, we will simply show how sample complexity is reduced based on $\omega$, $\alpha$, and $\zeta$. If we define all state-action pairs as the nodes of the feedback graph $\mathcal{G}$, $\mathcal{G}$ can be divided into two parts $\mathcal{G}_1$ and $\mathcal{G}_2$, where $\mathcal{G}_1$ is a connected graph, $\mathcal{G}_2$ is a graph without any edge, and $\mathcal{G}=\mathcal{G}_1\cup \mathcal{G}_2$. 

For the uncensored case, $\mathcal{G}=\mathcal{G}_1$ is a complete graph and $\mathcal{G}_2=\emptyset$. The complete feedback graph has a good property which is $\omega=\alpha=\zeta=1\ll |\mathcal{S}||\mathcal{A}|$. For the censored case, $\mathcal{G}$ consists of $\mathcal{G}_1$ for $\hat{y}_t\leq y_t$ and $\mathcal{G}_2$ for $\hat{y}_t> y_t$. Thus the feedback graph's property should be $\alpha=\zeta=(y^{\text{max}}-y_t)\times(a^{\text{max}})^L+1(y_t\neq0)$ and $\omega=(y^{\text{max}}-y_t)\times(a^{\text{max}})^L+y_t$. We can obtain that $\alpha=\zeta\leq\omega\leq|\mathcal{S}||\mathcal{A}|$, where $\alpha=\zeta=\omega$ holds iff $y_t=0\&1$ and $\omega=|\mathcal{S}||\mathcal{A}|$ holds iff $y_t=0$. Overall, sample complexity after using FG can be reduced.

\section{Theoretical Analysis}\label{Appendix:Theoretical Analysis}
Let us analyze the sample complexity and update probability scenario by scenario from the simplest case to real case. Let us start with \textbf{scenario 2} since \textbf{scenario 1} is analyzed in section \ref{sect:theoretical}.

\textbf{Scenario 2:} Consider a graph $\mathcal{G}$ with all state-action pairs as the nodes. Assume all nodes, once one node is sampled, all of these nodes can be sampled and updated at the same time. Thus $G$ can be regarded as $\mathcal{G}=\mathcal{G}_1\cup \mathcal{G}_2$, where $\mathcal{G}_1$ is a complete graph and $\mathcal{G}_2=\emptyset$. This case is the same as Synchronous Q-Learning in \cite{li2024q}.   

\textbf{Lemma 2:} The sample complexity of the Q-learning with feedback graph under scenario 2 is $\Tilde{O}(\frac{1}{(1-\gamma)^4\epsilon^2})$ with learning rate being $(1-\gamma)^3\epsilon^2$.

\textbf{Scenario 3:} Consider a graph $\mathcal{G}$ with all state-action pairs as the nodes. Assume for some nodes, once one node is sampled, all of these nodes can be sampled and updated at the same time. For other nodes, when one node is sampled, only itself can be updated. Thus $\mathcal{G}$ can be regarded as $\mathcal{G}=\mathcal{G}_1\cup \mathcal{G}_2$, where $\mathcal{G}_1$ is a complete graph and $\mathcal{G}_2$ consists of nodes without any edge.  

\textbf{Lemma 3:} The sample complexity of the Q-learning with FG under scenario 3 is $\Tilde{O}(\frac{1}{\Tilde\mu_{\text{min}}(1-\gamma)^5\epsilon^2}+\frac{t_\text{mix}}{\Tilde\mu_{\text{min}}(1-\gamma)})$, where $\Tilde\mu_{\text{min}}=\min[\min_{(\boldsymbol{s},a)\in G_2}\mu(\boldsymbol{s},a),\sum_{(\boldsymbol{s},a)\in \mathcal{G}_1}\mu(\boldsymbol{s},a)]$

Scenario 1 indicates the IC environment without feedback graph and scenario 2 indicates the uncensored case of the IC environment with feedback graph. As for the censored case of the IC environment, we simplify it in scenario 3 by assuming $\mathcal{G}_1$ is a complete graph. We can see that the sample complexity order is $\Tilde{O}(\frac{1}{\mu_{\text{min}}(1-\gamma)^5\epsilon^2}+\frac{t_\text{mix}}{\mu_{\text{min}}(1-\gamma)})\geq\Tilde{O}(\frac{1}{\Tilde\mu_{\text{min}}(1-\gamma)^5\epsilon^2}+\frac{t_\text{mix}}{\Tilde\mu_{\text{min}}(1-\gamma)})>\Tilde{O}(\frac{1}{(1-\gamma)^4\epsilon^2})$, since $\min_{(\boldsymbol{s},a)\in \mathcal{G}}\mu(\boldsymbol{s},a)<\min[\min_{(\boldsymbol{s},a)\in G_2}\mu(\boldsymbol{s},a),\sum_{(\boldsymbol{s},a)\in \mathcal{G}_1}\mu(\boldsymbol{s},a)]<1$, where $1$ indicates $\mu_{\text{min}}=1$ in scenario 2.

Now, Let us loosen the assumption that $\mathcal{G}_1$ is a complete graph in scenario 3.

\textbf{Scenario 4:} Consider a graph $G=G_1\cup G_2$ with all state-action pairs as the nodes. We define $v_i<v_j$ and $s_i<s_j$ if $s_i[0]<s_j[0]$. For nodes in $G_1$, once one node $v_i$ is sampled, $\forall v_j<v_i$ can be sampled and updated at the same time. For nodes in $G_2$, when one node is sampled, only itself can be updated. Thus $G$ can be regarded as $G=G_1\cup G_2$, where $G_1$ is a connected graph and $G_2=\emptyset$.

\textbf{Lemma 4:} The sample complexity of the Q-learning with feedback graph under Assumption 4 is $\Tilde{O}(\frac{1}{\hat\mu'_{min}(1-\gamma)^5\epsilon^2}+\frac{t_{mix}}{\hat\mu'_{min}(1-\gamma)})$, where $\hat\mu'_{min}=\min[\min_{(s,a)\in G_2}\mu(s,a),\sum_{\{s,a|s\in G_1;\forall \hat{s}\in G_1, s>\hat{s}\}}\mu(s,a)]$.

Based on the above scenarios and lemmas, let us consider the lost-sales IC environment with constant demand $d$.

\textbf{Scenario 5:} Consider a graph $\mathcal{G}$ with all state-action pairs as the nodes.  Assume for nodes with $y\geq d$, once one node is sampled, all of these nodes can be sampled and updated simultaneously. For nodes with $y<d$, only nodes with $y'\leq y$ can be sampled and updated simultaneously. Thus $\mathcal{G}$ can be regarded as $\mathcal{G}=\mathcal{G}_1\cup \mathcal{G}_2$. 


\begin{table*}[t!]
    \begin{center}
    \caption{Parameters of Rainbow-FG and TD3-FG.}
    \label{para-rainbow-fg}
    \begin{tabular}{ c c c c }
    \hline
     Parameter &Value &Parameter &Value\\
     \hline
     Training Episodes     & 100      & Training Steps/Episode        &1000    \\
     Testing Frequency     & 1        & Testing Steps/Episode         &400    \\
     Batch Size            & 128      & Batch Size for FG           &256    \\
     Replay Buffer Size    & 12000    & Replay Buffer Size for FG   &192000    \\
     $\epsilon$            & 0.1      & $\gamma$ & 0.995    \\
     lr                    &1e-4      & Target Update Frequency  & 100  \\
     $V^{\text{min}}$             & -200     &  $V^{\text{max}}$      &0    \\
     $N^{atom}$            &51       & Hidden Layer Size & 512\\
     Intrinsic Reward Weight  &0.01       & Intrinsic Reward Discount Factor & 0.9\\
    \hline
    \end{tabular}
    \end{center}
\end{table*}

\textbf{Lemma 5:} The sample complexity of the Q-learning with feedback graph under scenario 5 is $\Tilde{O}(\frac{1}{\Tilde\mu_{\text{min}}(1-\gamma)^5\epsilon^2}+\frac{t_\text{mix}}{\Tilde\mu_{\text{min}}(1-\gamma)})$, where $\Tilde\mu_{\text{min}}=\sum_{(s,a)\in \mathcal{G}; y\geq d}\mu(s,a)$.

\textbf{Proof:}

For $\{(s,a)|(s,a)\in \mathcal{G}; y\geq d\}$, we have $\Tilde\mu_{\text{min}}^{y\geq d}=\sum_{(s,a)\in \mathcal{G}; y\geq d}\mu(s,a)$. 

For $\{(s,a)|(s,a)\in \mathcal{G}; y< d\}$, we have $\Tilde\mu_{\text{min}}^{y<d}=\sum_{y=d}\mu(s,a)+\mu_{\text{min}}^{y\geq d}$. 

Thus we have $\Tilde\mu_{\text{min}}=\Tilde\mu_{\text{min}}^{y\geq d}$. 


If $\mu_{\text{min}}$ appears in $\{(s,a)|(s,a)\in \mathcal{G}; y\geq d\}$, we have $\Tilde\mu_{\text{min}}=\Tilde\mu_{\text{min}}^{y\geq d}=\sum_{(s,a)\in \mathcal{G}; y\geq d}\mu(s,a)>(y^{\text{max}}-d)|A|^L\min_{(s,a)\in \mathcal{G}; y\geq d}\mu(s,a)=(y^{\text{max}}-d)|A|^L\mu_{\text{min}}$.

If $\mu_{\text{min}}$ appears in $\{(s,a)|(s,a)\in \mathcal{G}; y<d\}$, which means $\mu_{\text{min}}<\min_{(s,a)\in \mathcal{G}; y\geq d}\mu(s,a)$, we have $\Tilde\mu_{\text{min}}=\Tilde\mu_{\text{min}}^{y\geq d}=\sum_{(s,a)\in \mathcal{G}; y\geq d}\mu(s,a)>(y^{\text{max}}-d)|A|^L\min_{(s,a)\in \mathcal{G}; y\geq d}\mu(s,a)>(y^{\text{max}}-d)|A|^L\mu_{\text{min}}$.

\textbf{Thus $\Tilde\mu_{\text{min}}$ under scenario 5 improves at least $(y^{\text{max}}-d)|A|^L$ times than that under scenario 1.}

\textbf{Proof done.}

Now we lossen the assumption of the constant demand $d$ to $d_t\sim P_d(d)$.

\textbf{Scenario 6:} Consider a graph $\mathcal{G}$ with all state-action pairs as the nodes.  Consider all nodes satisfying $y\geq d_t$, if one is sampled by the policy, all can be sampled by the FG module and updated simultaneously. For nodes with $y<d_t$, only nodes with $y'\leq y$ can be sampled by the FG module and updated simultaneously. Thus $\mathcal{G}$ can be regarded as $\mathcal{G}=\mathcal{G}_1\cup \mathcal{G}_2$.

\textbf{Proof of Theorem 1:}

For each $(\boldsymbol{s},a)$ and each possible value of $d$, we have 
\begin{equation}
\begin{aligned}
    \Tilde\mu^{y\geq d}(\boldsymbol{s},a|d)=\sum_{\substack{(\boldsymbol{\hat s},\hat a)\in \mathcal{G}\\ \hat y\geq d}}\mu(\boldsymbol{\hat s},\hat a|d)\geq\mu^{y\geq d}(\boldsymbol{s},a|d),\ \{(\boldsymbol{s},a)|(\boldsymbol{s},a)\in \mathcal{G}; y\geq d\}.
\end{aligned}
\end{equation}

\begin{equation}
\begin{aligned}
    \Tilde\mu^{y<d}(\boldsymbol{s},a|d)&=\sum_{\substack{(\boldsymbol{\hat s},\hat a)\in \mathcal{G}\\y\leq\hat y\leq d}}\mu(\boldsymbol{\hat s},\hat a|d)+ \sum_{\substack{(\boldsymbol{\hat s},\hat a)\in \mathcal{G}\\ \hat y\geq d}}\mu(\boldsymbol{\hat s},\hat a|d)\\
    &\geq\mu^{y\leq d}(\boldsymbol{s},a|d),\ \{(\boldsymbol{s},a)|(\boldsymbol{s},a)\in \mathcal{G}; y< d\}.
\end{aligned}
\end{equation}

Thus we have
\begin{equation}
    \Tilde\mu(\boldsymbol{s},a)=E_{d\sim P_d}[\Tilde\mu(\boldsymbol{s},a|d)] \geq E_{d\sim P_d}[\mu(\boldsymbol{s},a|d)]=\mu(\boldsymbol{s},a).
\end{equation}

Then we formulate $\Tilde\mu(\boldsymbol{s},a)$ in details:

\begin{equation}
\begin{split}
\Tilde\mu(\boldsymbol{s},a)
&=E_{d\sim P_d}[\Tilde\mu(\boldsymbol{s},a|d)] \\ 
&= \sum_{d=y}^{d^{\text{max}}}P_d(d)\sum_{\substack{(\boldsymbol{\hat s},\hat a)\in \mathcal{G}\\ \hat y\geq d}}\mu(\boldsymbol{\hat s},\hat a|d) +\sum_{d=0}^{y-1}P_d(d)\sum_{\substack{(\boldsymbol{\hat s},\hat a)\in \mathcal{G}\\ y\leq\hat y\leq d}}\mu(\boldsymbol{\hat s},\hat a|d)+\sum_{\substack{(\boldsymbol{\hat s},\hat a)\in \mathcal{G}\\ \hat y\geq d}}\mu(\boldsymbol{\hat s},\hat a|d)\\ 
&=\sum_{d=0}^{d^{\text{max}}}P_d(d)\sum_{\substack{( \boldsymbol{\hat s},\hat a)\in \mathcal{G}\\ \hat y\geq d}}\mu(\boldsymbol{\hat s},\hat a|d) + \sum_{d=0}^{y-1}P_d(d)\sum_{\substack{(\boldsymbol{\hat s},\hat a)\in \mathcal{G}\\ y\leq\hat y\leq d}}\mu(\boldsymbol{\hat s},\hat a|d).
\end{split}
\end{equation}
\textbf{Proof done.}

\section{Parameters of Experiments}\label{appendix:para-of-rainbow-fg}
Table \ref{para-rainbow-fg} shows the parameters for Rainbow-FG and TD3-FG. Table \ref{table:para-of-env} shows the parameters for the single-item environment. As for multi-item environment, we set $P=[4,9]$ and $L=[4,3]$ for 2 items, $P=[4,9,19]$ and $L=[4,3,2]$ for 3 items, $P=[4,9,4,9,19]$ and $L=[4,4,3,3,2]$ for 5 items. As for multi-echelon environment, we set the warehouse with $P=4$ and $L=4$; $P=4$ and $L=4$ for 1 retailer, $P=[4,9]$ and $L=[4,4]$ for 2 retailers, $P=[4,9,4,9]$ and $L=[4,4,3,3]$ for 4 retailers. Other parameters are kept the same as the single-item environment.

\begin{table}[htb] 
    \begin{center}
    \caption{Parameters of the environment. $L$ and $p$ are two important parameters where $L$ reflects the dimensionality of the state and $p$ reflects the seriousness of lost sales.}
    \setlength{\tabcolsep}{1.5mm}
    \label{table:para-of-env}
    \renewcommand{\arraystretch}{1.3}
    \begin{tabular}{ cllllllll}
    \hline
    Parameter &$L$ &$p$ &$y^{\text{max}}$ &$a^{\text{max}}$ &$c$ &$h$ &$d^m$ &$d^{\text{max}}$\\ 
    \hline
    Value & $4$  & $4$  & $100$  &$20$ & $0$ & $1$ & $5$ &$20$\\
    \hline
    \end{tabular}
    \end{center}
\end{table}

\section{Compute Resources}\label{appendix:comp-resou}
Experiments are carried out on Intel (R) Xeon (R) Platinum 8375C CPU @ 2.90GHz and NVIDIA GeForce RTX 3080 GPUs. All the experiments can be done within one day.

\section{More experiments for FG}\label{appendix:more-exp-fg}

\subsection{Sample Efficiency Comparison for Intrinsic Reward}
This section investigates the effect of intrinsic reward designed for the IC problem. Figure \ref{exp:inr} illustrates the learning process of Rainbow-FG with and without the intrinsic reward, referred to as Rainbow-FG w/ inr and Rainbow-FG w/o inr, respectively. The results demonstrate that the designed intrinsic reward significantly improves sample efficiency during the initial stages of training in the environment with $p=4, L=4$, and $d^m=5$. However, as the learning process reaches the Constant Order level, the intrinsic reward does not exhibit a substantial impact. We attribute it to the simplicity of the environment, which weakens the effect of the intrinsic reward. To further evaluate its effectiveness, we conduct experiments in more complex settings with $p=19, L=8$, and varying values of $d^m=$ (5, 10, 15). In these environments, we observe a clear effect of the intrinsic reward during the whole training process. Moreover, Rainbow-FG w/ inr demonstrates a more stable training process compared to Rainbow-FG w/o inr, achieving slightly lower cost.

\begin{figure}[h!]
    \centering   
    \subfigure[p=4, L=4, $d^m$=5] 
    {   \hspace{-5mm}
        \begin{minipage}[h]{.22\textwidth } 
    		\centering
    		\includegraphics[scale=0.35]{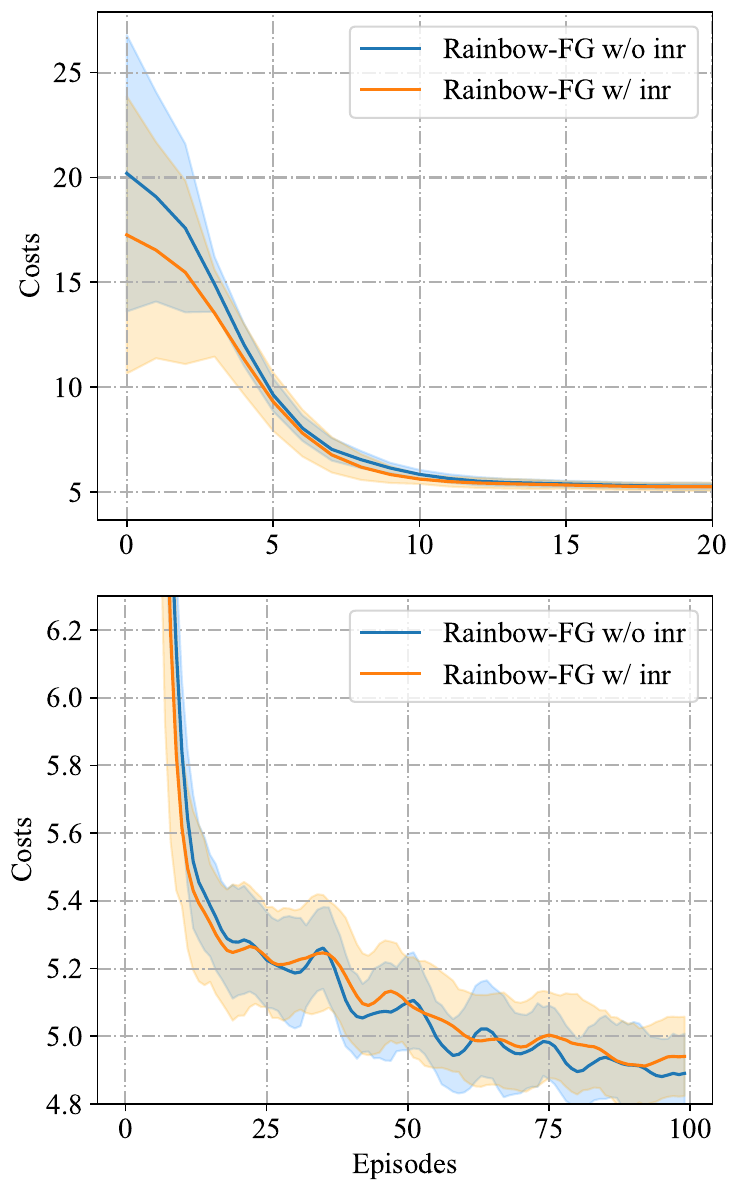}
        \end{minipage}
    }
    \subfigure[p=19, L=8, $d^m$=5]
    {
        \begin{minipage}[h]{.22\textwidth }
    		\centering
    		\includegraphics[scale=0.35]{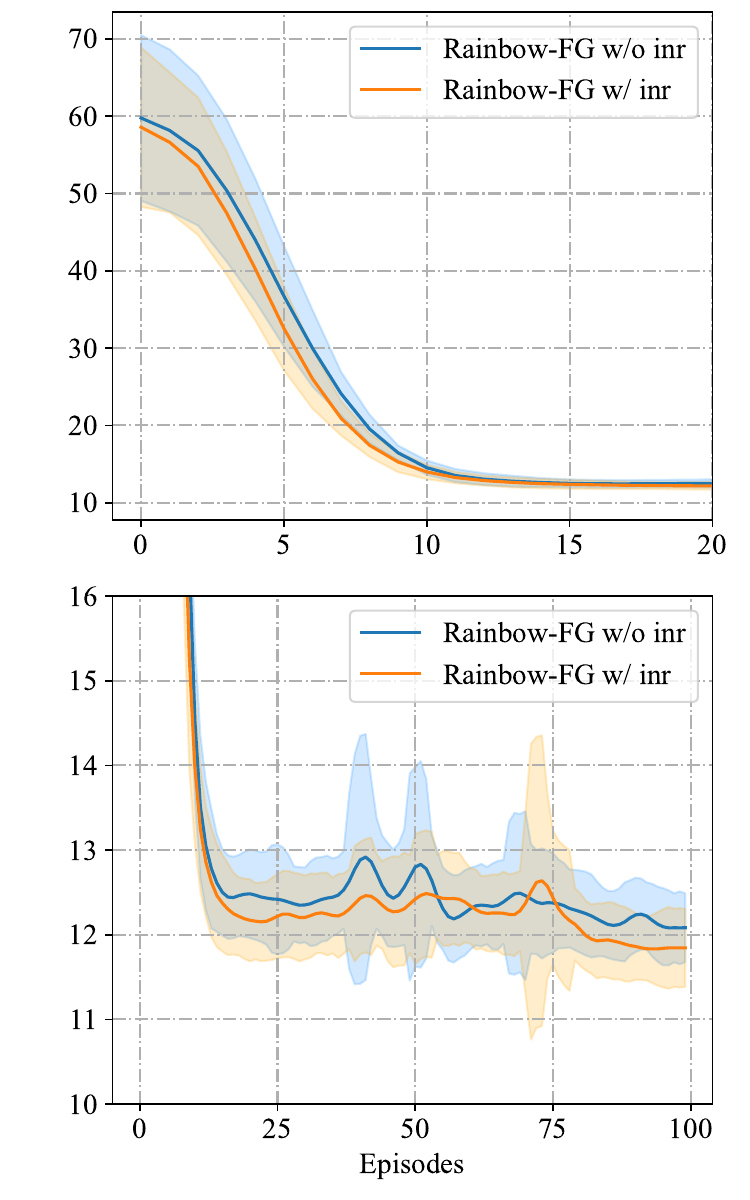}
        \end{minipage}
    }
    \subfigure[p=19, L=8, $d^m$=10]
    {
        \hspace{-5mm}
        \begin{minipage}[h]{.22\textwidth }
    		\centering
    		\includegraphics[scale=0.35]{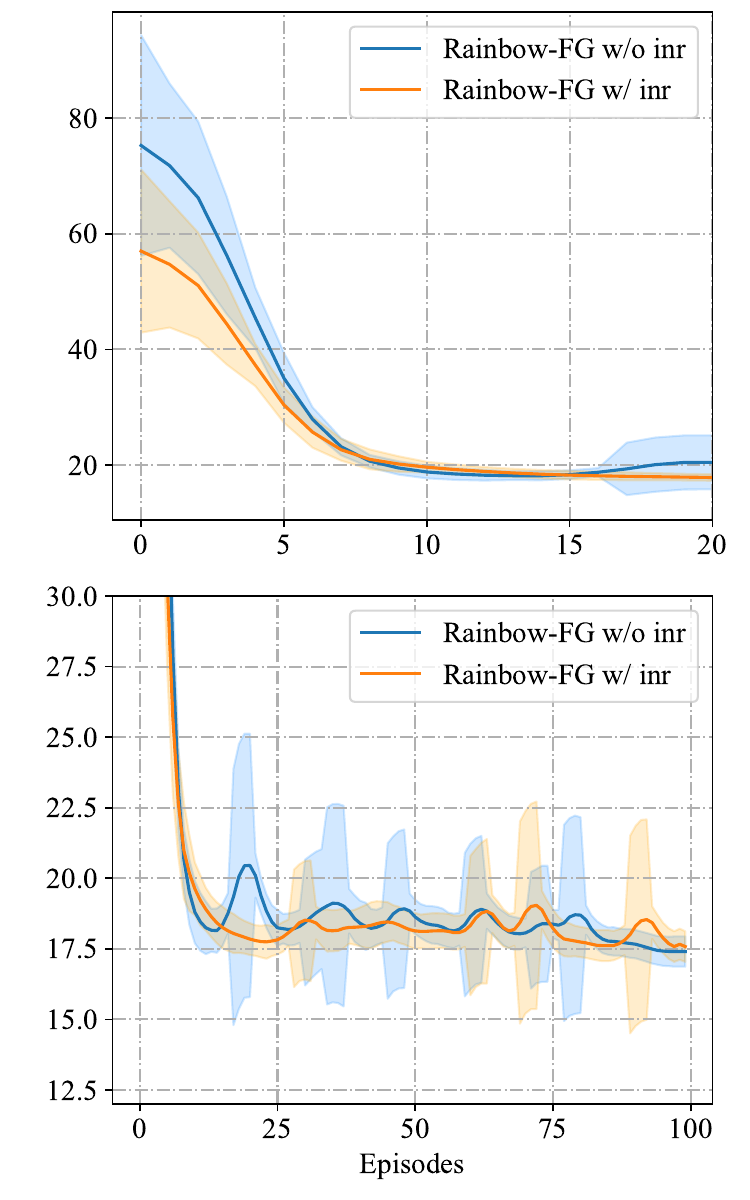}
        \end{minipage}
    }
    \subfigure[p=19, L=8, $d^m$=15]
    {
        \begin{minipage}[h]{.22\textwidth }
    		\centering
    		\includegraphics[scale=0.35]{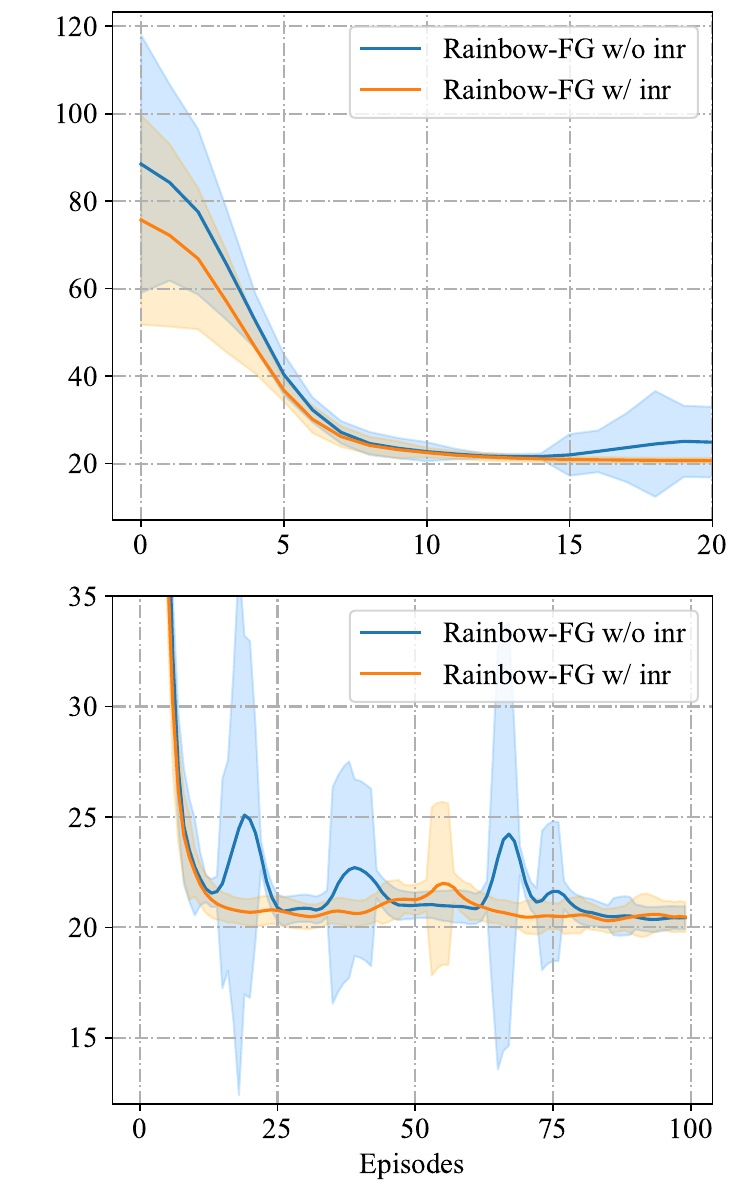}
        \end{minipage}
    }
    \caption{Learning process of Rainbow-FG with and without intrinsic reward. The two rows have the same meaning with Figure 2.}
    \label{exp:inr}
\end{figure}

\section{Hyperparameter Analysis}\label{appendix:hyper-analy}

\subsection{Sample Efficiency under Different Exploration Level}
Figure \ref{fig:more-exp-fg} shows the learning process of Rainbow and Rainbow-FG with different exploration parameters in more settings. The learning processes in these settings show similar properties in section \ref{subsec:exp-fg-single-item}.

\subsection{Feedback Graph Size}
Theoretically, we should get enough side experiences by considering every term of $S$ and $A$ to get the largest size of $G_1$. However, in practice, the time or resources may be limited so that only part of the side information can be obtained. Based on this question, this section focuses on the effect of different sizes of the feedback graph.

The default setting only constructs the feedback graph considering the current inventory $y_t$, which is the first term of $\boldsymbol{s}_t$, and action $a_t$.  Each comparison group adds one following term of $\boldsymbol{s}_t$ when constructing the feedback graph. Figure \ref{fig_exp3} illustrates the learning process of Rainbow-FG with different sizes of FG. The result shows that the sample efficiency is more sensitive to the size of FG at the initial stage than at the final stage. As the size of FG increases, the improvement of the sample efficiency mainly occurs at the initial stage. At the final stage, Rainbow-FG w/ s[0:1] \& a and s[0:4] \& a first reach the final level and then follows Rainbow-FG with s[0:3] \& a.

\subsection{Intrinsic Reward Weight}
To test the sensitivity of our intrinsic reward design, we test our method with different intrinsic reward weights. Figure \ref{fig_exp5} shows the detailed learning process. A larger intrinsic reward weight tends to have higher sample efficiency during the initial stages of training. This result demonstrates the benefits of improving sample efficiency for the intrinsic reward method. However, large intrinsic reward weights can affect the performance of the final stage. The main reason is that adding the intrinsic reward to the extrinsic reward changes the original objective to be optimized. 

\begin{figure}
\centering
\subfigure[$p=4,L=3$]{
\hspace{-5mm}
\includegraphics[width=0.5\textwidth]{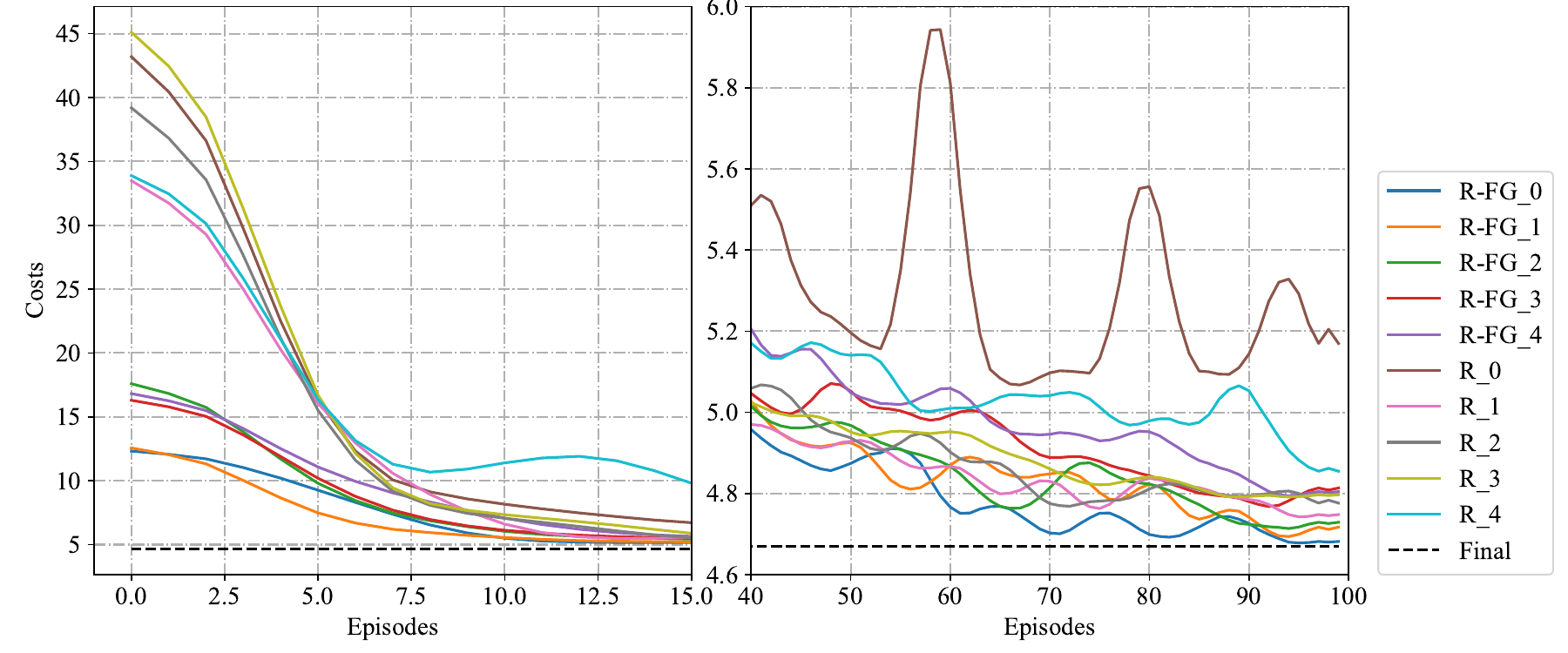} 
}
\subfigure[$p=9,L=4$]{
\hspace{-5mm}
\includegraphics[width=0.5\textwidth]{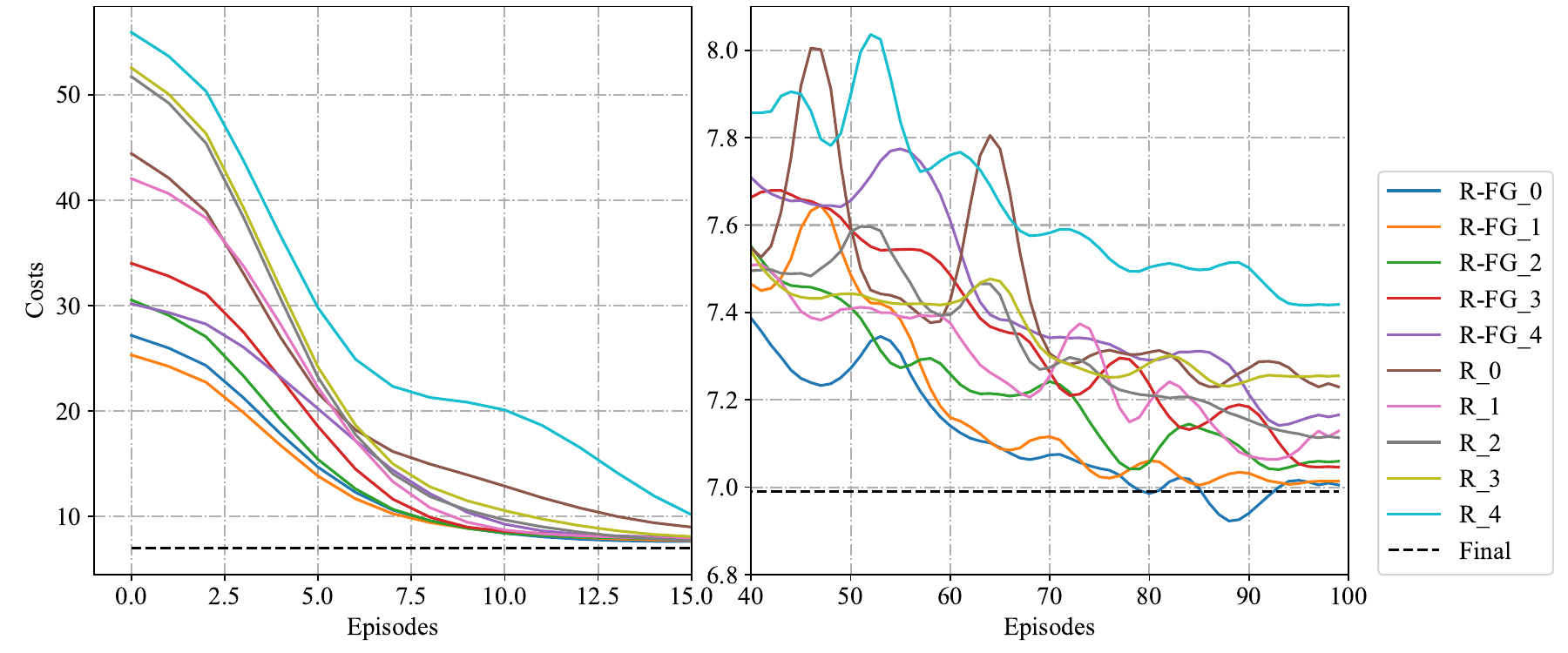} 
}
\subfigure[$p=9,L=3$]{
\hspace{-5mm}
\includegraphics[width=0.5\textwidth]{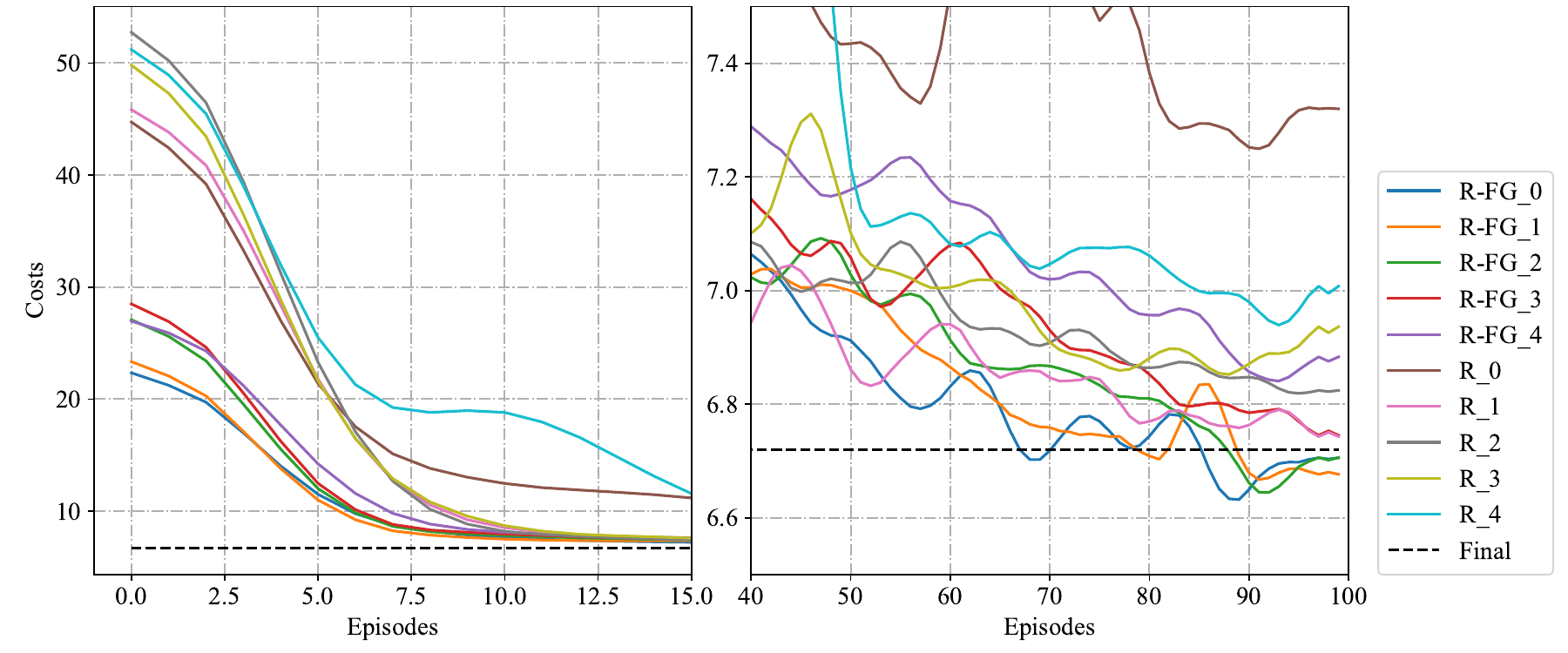} 
}
\subfigure[$p=9,L=2$]{
\hspace{-5mm}
\includegraphics[width=0.5\textwidth]{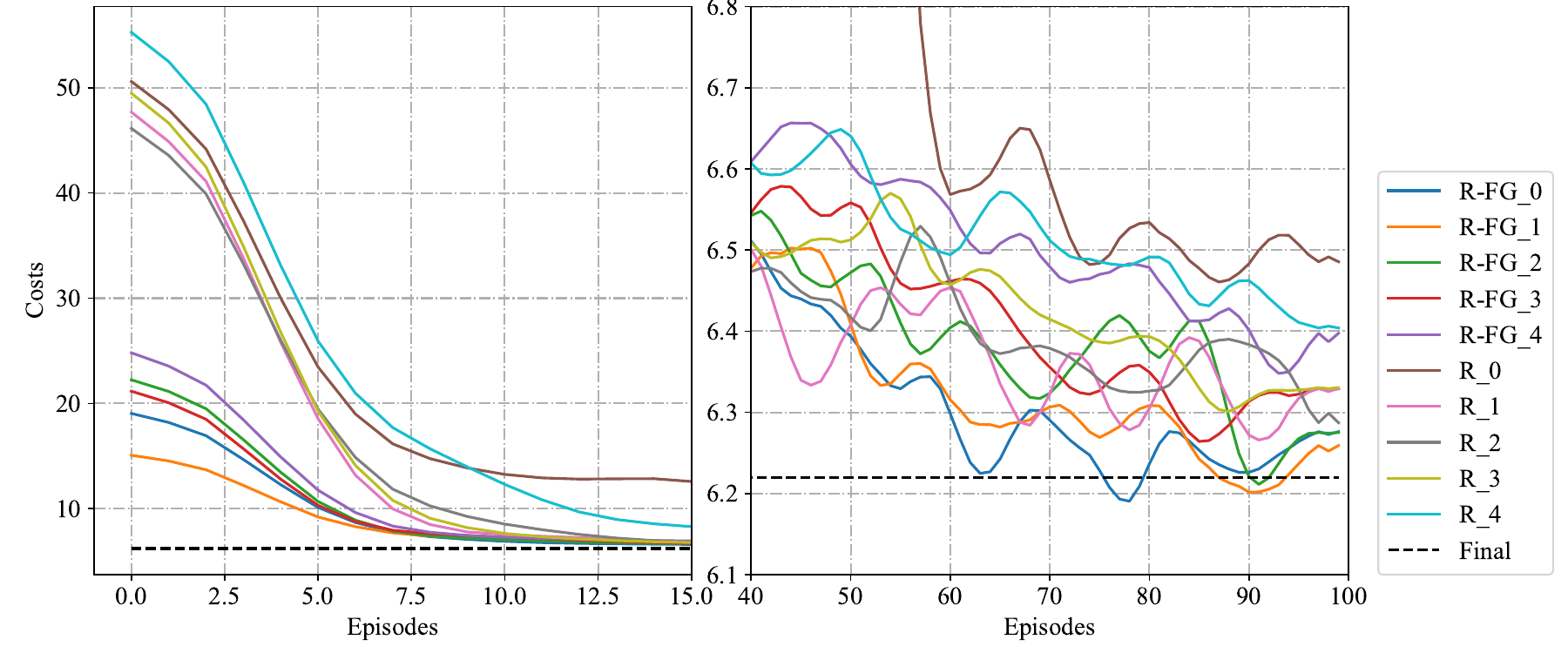} 
}
\caption{Learning process of Rainbow and Rainbow-FG under the IC environment. ``R\_x'' denotes Rainbow with $\epsilon=0.x$ and ``R-FG\_x'' denotes Rainbow-FG with $\epsilon=0.x$. ``Final'' denotes the optimal result of Rainbow-FG. The first column is the initial-stage view and the second column is the near-convergence view of the learning process.}
\label{fig:more-exp-fg}
\end{figure}

\newpage

\begin{figure*}[t]
\centering 
\includegraphics[width=1.03\textwidth]{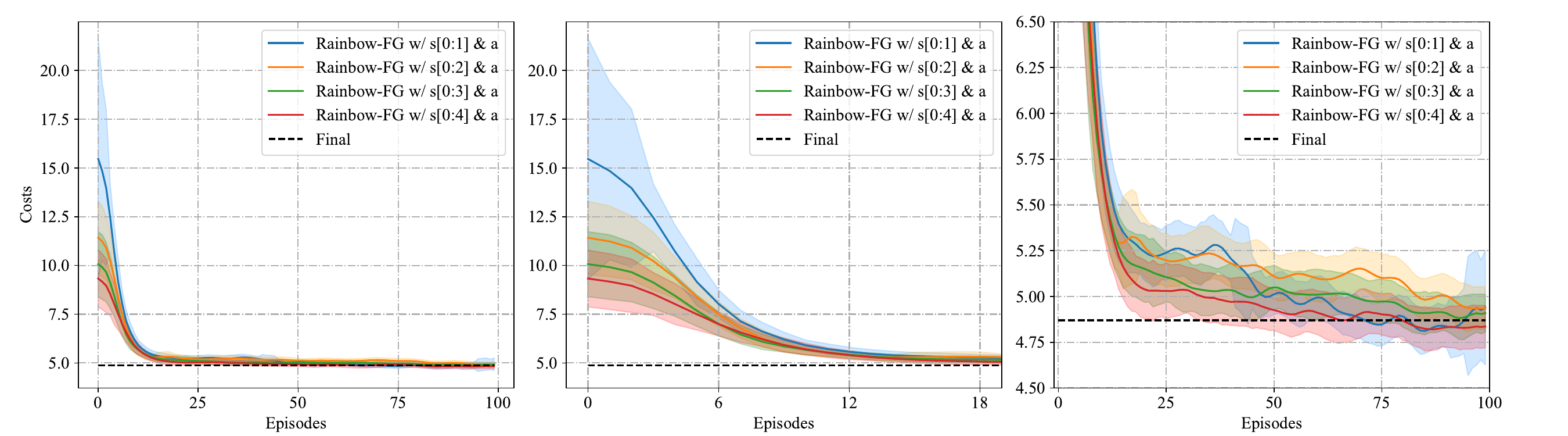} 
\caption{Learning process of Rainbow-FG with different feedback graph size. The first column is the full view of the learning process. The second and third columns are different partial views of the first column.} 
\label{fig_exp3} 
\end{figure*}

\begin{figure*}[t]
\centering 
\includegraphics[width=1\textwidth]{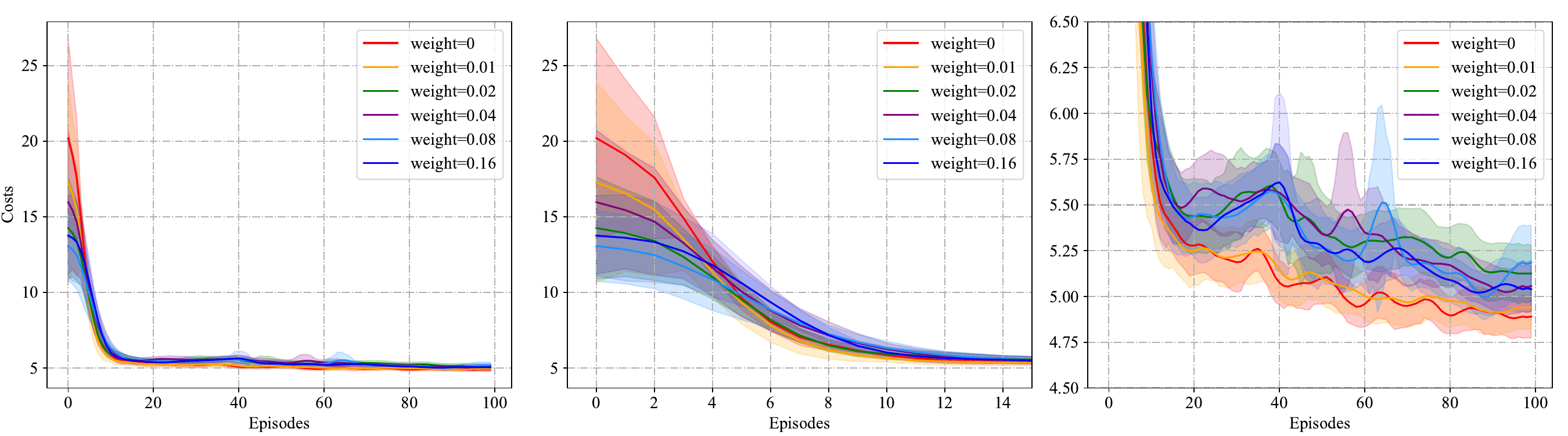} 
\caption{Learning process of Rainbow-FG with different intrinsic reward weights. The intrinsic reward weight is multiplied by 2 each time. The first column is the full view of the learning process. The second and third columns are different partial views of the first column.} 
\label{fig_exp5} 
\end{figure*}

\end{APPENDICES}

\end{document}